\documentclass[conference]{IEEEtran}
\IEEEoverridecommandlockouts
\usepackage{cite}
\usepackage{amsmath,amssymb,amsfonts}
\usepackage{algorithmic}
\usepackage{graphicx}
\usepackage{textcomp}
\usepackage{xcolor}
\usepackage{times}
\usepackage{epsfig}
\usepackage{subfigure}
\usepackage{upgreek}
\usepackage{booktabs}
\usepackage{colortbl}
\usepackage[ruled,linesnumbered]{algorithm2e}
\def\BibTeX{{\rm B\kern-.05em{\sc i\kern-.025em b}\kern-.08em
    T\kern-.1667em\lower.7ex\hbox{E}\kern-.125emX}}
\begin{document}

\makeatletter
\DeclareRobustCommand\onedot{\futurelet\@let@token\@onedot}
\def\@onedot{\ifx\@let@token.\else.\null\fi\xspace}

\def\eg{\emph{e.g}\onedot} \def\Eg{\emph{E.g}\onedot}
\def\ie{\emph{i.e}\onedot} \def\Ie{\emph{I.e}\onedot}
\def\cf{\emph{c.f}\onedot} \def\Cf{\emph{C.f}\onedot}
\def\etc{\emph{etc}\onedot} \def\vs{\emph{vs}\onedot}
\def\wrt{w.r.t\onedot} \def\dof{d.o.f\onedot}
\def\etal{\emph{et al}\onedot}
\makeatother

\title{Self-distilled Knowledge Delegator for Exemplar-free Class Incremental Learning}

\author{\IEEEauthorblockN{Fanfan Ye, Liang Ma, Qiaoyong Zhong\textsuperscript{$\star$}, Di Xie, Shiliang Pu}
\IEEEauthorblockA{\textit{Hikvision Research Institute} \\
% Hangzhou, China \\
\{yefanfan,maliang6,zhongqiaoyong,xiedi,pushiliang.hri\}@hikvision.com}
}

\maketitle 

{\let\thefootnote\relax\footnotetext{$^\star$Corresponding author.}}

\begin{abstract}
  Exemplar-free incremental learning is extremely challenging due to inaccessibility of data from old tasks. In this paper, we attempt to exploit the knowledge encoded in a previously trained classification model to handle the catastrophic forgetting problem in continual learning. Specifically, we introduce a so-called knowledge delegator, which is capable of transferring knowledge from the trained model to a randomly re-initialized new model by generating informative samples. Given the previous model only, the delegator is effectively learned using a self-distillation mechanism in a data-free manner. The knowledge extracted by the delegator is then utilized to maintain the performance of the model on old tasks in incremental learning. This simple incremental learning framework surpasses existing exemplar-free methods by a large margin on four widely used class incremental benchmarks, namely CIFAR-100, ImageNet-Subset, Caltech-101 and Flowers-102. Notably, we achieve comparable performance to some exemplar-based methods without accessing any exemplars.
\end{abstract}


%%%%%%%%% BODY TEXT
\section{Introduction}

Class Incremental Learning (CIL) is a prevalent and nontrivial problem in the community. It requires Deep Neural Networks (DNNs) to have the ability to handle streaming tasks instead of only one task. Due to the notorious catastrophic forgetting phenomenon~\cite{goodfellow2013empirical, Hou2019}, DNNs always suffer a severe performance degradation on the previously learned tasks while learning a new task. 

Many excellent works have been proposed to alleviate the catastrophic forgetting for CIL. A common and straightforward way is to store some exemplars from the old tasks and replay them in the following incremental learning~\cite{castro2018end, Hou2019, Liu2020, Rebuffi2017, tao2020topology, wu2019large, he2021tale} (referred to as exemplar-based methods), which have achieved remarkable performance. While they are admirable, in practice, this is not the case. As privacy protection becomes more stringent, the training data from old tasks are often strictly unavailable, which means the exemplars are unavailable. This limitation makes CIL more challenging. Concerning the case, some regularization-based approaches are first proposed, such as MAS~\cite{Aljundi2018}, EWC~\cite{Kirkpatrick2017}, LwF~\cite{Li2018} and SI~\cite{Zenke2017} (referred to as exemplar-free methods). However, the performances of these methods are poor, especially as the number of tasks increases. 
\begin{figure}[t]
\centering
\includegraphics[width=1\columnwidth]{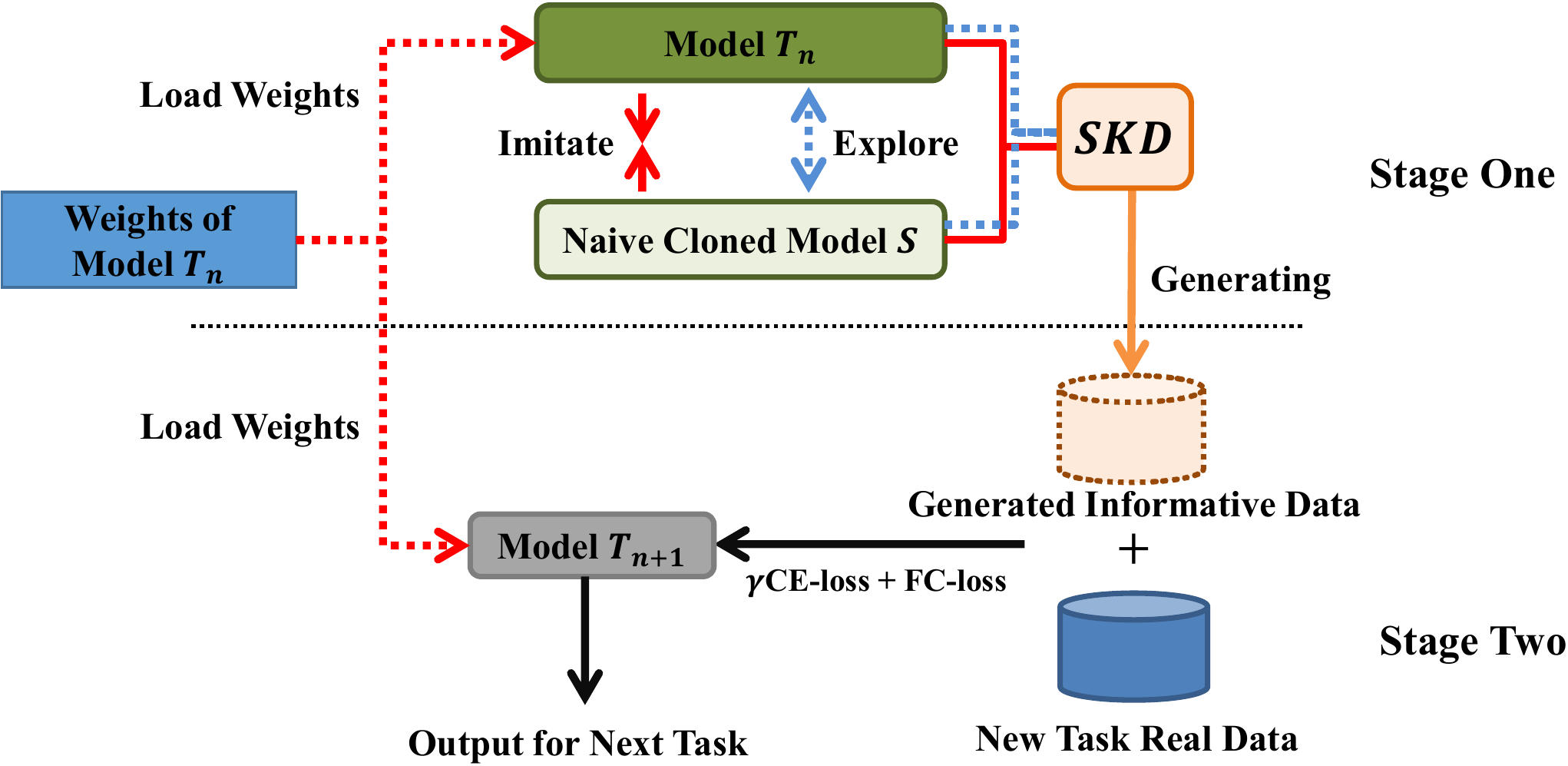}
  \caption{An overview of the proposed exemplar-free class incremental learning framework. In stage one, the Self-distilled Knowledge Delegator (SKD) is trained by a novel imitate \& explore adversarial strategy. In stage two, informative data generated by SKD and real data of the new task are combined for continual learning. ``Load Weights'' means to load pre-trained parameters.}
\label{pipline}
\end{figure}
Intuitively, for CIL, one of the essential practices is to preserve the feature space of old tasks. We notice that this motivation is similar to the problem of data-free knowledge transfer~\cite{liu2021data, Chen2019, Fang2019, Haroush2020, Micaelli2019, Ye2020, Yin2020}. These works have proved that the generative networks can be considered a potential intermediate messenger, which transfers knowledge and keeps the feature space constant without touching any real data. These promising attempts shed light on the design of exemplar-free CIL methods.
%Naturally, a question comes to our mind. Can we replay the pseudo data in incremental learning and alleviate the catastrophic forgetting? 
Nevertheless, the methods mentioned above cannot be directly used in CIL since they can only work well on some very simple datasets with low resolution such as MNIST~\cite{lecun1998gradient} and CIFAR-10~\cite{krizhevsky2009learning}. To meet more complex settings, significant improvements are required.

In this paper, we propose a novel exemplar-free CIL framework from the perspective of data-free knowledge transfer. As illustrated in Fig.~\ref{pipline}, a so-called Self-distilled Knowledge Delegator (SKD) is introduced and serves as the key to CIL. SKD is trained to extract knowledge of a previously learned model. And the knowledge is used to alleviate catastrophic forgetting during incremental learning. Specifically, the whole framework consists of two stages in each CIL task. 1) An SKD model is trained using a self-distillation mechanism, which is totally data-free and only requires the pre-trained classification model ${T}$. A novel imitate \& explore adversarial training strategy is introduced to help SKD to synthesize more informative data. 2) Given the trained SKD model, pseudo data generated by SKD and real data of current task are combined for CIL. A feature consolidation loss (FC-loss) and a cross-entropy loss (CE-loss) are adopted. To verify the effectiveness of the proposed method, we conduct extensive experiments on four widely used CIL benchmarks, \ie CIFAR-100~\cite{krizhevsky2009learning}, ImageNet-Subset, Caltech-101~\cite{fei2004learning} and Flowers-102~\cite{nilsback2008automated}. Our method achieves state-of-the-art performance compared with existing exemplar-free methods, and is very comparable to some exemplar-based methods without accessing any exemplars.
Our main contributions can be summarized as follows.
\begin{itemize}
  \item We propose a novel exemplar-free class incremental learning framework based on the knowledge delegator, which is the first effective attempt that adopts a generative network for CIL in a totally data-free manner.
\item We introduce the self-distillation mechanism to train SKD effectively, which works well not only on simple and low-resolution datasets, but also high-resolution and fine-grained benchmarks.
\item Our method achieves state-of-the-art performance and outperforms classic exemplar-free methods by a large margin, even exceeding some excellent exemplar-based methods.
\end{itemize}

\section{Related Works}
\subsection{Class Incremental Learning}
Various methods have been proposed for CIL, which can be roughly divided into three categories as follows.

\textbf{Regularization-based methods} aim to preserve important neurons (with regard to old tasks) of the neural network by penalizing changes over these parameters while training on a new task. EWC~\cite{Kirkpatrick2017} is a fundamental method that directly applies parameter regularization during network updating, which uses the diagonal of the Fisher information matrix to identify important neurons. MAS~\cite{Aljundi2018} estimates the neuron importance by measuring the sensitivity of the network to small perturbations on parameters. SI~\cite{Zenke2017} uses the path integral over the optimization trajectory. Recently, Belouadah~\etal~\cite{belouadah2019il2m,belouadah2020initial} assumed that all weights of the model should be normalized across tasks and standardized the initial classifier weights. Different from per-parameter regularization, LwF~\cite{Li2018} adds regularization on top of the network output.

\textbf{Network-based methods.} Through network capacity expansion, one can easily tackle catastrophic forgetting by growing a sub-network for each task, either logically or physically~\cite{yoon2017lifelong}. To reduce the network expansion speed, Mallya~\etal~\cite{mallya2018piggyback, mallya2018packnet} made a separate mask for every incremental task. Serra \etal~\cite{serra2018overcoming} and Masana \etal~\cite{masana2020ternary} argued that the mask should be learned for activations rather than parameters. Recently, the embedding network has drawn much attention to handle catastrophic forgetting by training a robust model before incremental learning~\cite{Yu2020}.

\textbf{Rehearsal-based methods.} The most straightforward way for incremental learning is to store real data and replay them during continual learning. Researchers attempted to obtain better performance under a limited storage budget, such as iCaRL~\cite{Rebuffi2017}, EEIL \cite{castro2018end}, BiC\cite{wu2019large}, LUCIR~\cite{Hou2019}, Mnemonics~\cite{Liu2020} and TPCIL~\cite{tao2020topology}. Others introduced a generator or variational auto-encoder to produce pseudo data or feature, referred to as pseudo-rehearsal~\cite{lesort2019generative}. \cite{shin2017continual} proposed an unconditional GAN while~\cite{Wu2018} changed the GAN from unconditional to conditional. Kemker \etal~\cite{kemker2017fearnet}, Xiang \etal~\cite{xiang2019incremental} and Liu \etal~\cite{Liu2020b} proposed different approaches to generate features rather than images. These pseudo-rehearsal methods still require access to real data when training the generator or encoder.

\subsection{Data-free Training}
Recently, several data-free training methods have arisen in the field of neural network compression, which show promising direction for continual learning. Chen \etal proposed DAFL~\cite{Chen2019} by bringing GAN to teacher-student knowledge distillation, in which the teacher network was regarded as the discriminator. Fang~\etal~\cite{Fang2019} and Micaelli~\etal~\cite{Micaelli2019} proposed similar strategy to make the generative network explore hard samples during adversarial training. However, these methods~\cite{Chen2019, Fang2019, Micaelli2019} can only work well on simple and low-resolution datasets such as MNIST and CIFAR-100, and fail on high-resolution and fine-grained datasets. Inspired by DeepDream\footnote{https://github.com/google/deepdream\label{deepdream}}, Haroush~\etal~\cite{Haroush2020} and Yin~\etal~\cite{Yin2020} proposed to invert images from a pre-trained model, demonstrating their capability on large-scale datasets like ImageNet~\cite{deng2009imagenet}. Image inversion methods need to store a huge amount of inverted data which is not acceptable for continual learning. Nevertheless, aforementioned data-free training methods open the door to continual learning in absence of data from previous tasks.

\section{Method}
In this section, we first briefly review the CIL task. Then we illustrate the details of the self-distillation mechanism to train the proposed SKD. After that, we present the overall exemplar-free class incremental learning process based on SKD.
\subsection{Preliminaries}
\textbf{Definition of Class Incremental Learning:} Class incremental learning as proposed in~\cite{Rebuffi2017}, aims to evaluate a classification model with a sequence of datasets $\{\mathcal{D}_0, \mathcal{D}_1, \dots, \mathcal{D}_N\}$ in an incremental manner. Formally, given the previous model ${T_{n}}$ trained on dataset ${\mathcal{D}_{n}}$, our goal is to learn a new classification model ${T_{n+1}}$ for both classes ${\mathcal{C}_{0\sim n}}$ and ${\mathcal{C}_{n+1}}$ only based on the new dataset ${\mathcal{D}_{n+1}}$, where $\mathcal{C}_n$ is the set of classes contained in $\mathcal{D}_n$. It is worth noting that ${\mathcal{C}_m} \cap \mathcal{C}_n = \emptyset, m\ne n$, which means there is no overlapping classes among the tasks. In this paper, we study the CIL problem under the constraint that data of previous tasks cannot be accessed or memorized by any means, which is more challenging than the previous exemplar-based setting. After incremental learning, the model ${T_{n+1}}$ will be evaluated on the validation set of all tasks that have been learned.

\textbf{Motivation of Our Work:} After investigating existing works, we find that the underlying rationale of alleviating catastrophic forgetting is to maintain the feature extraction capability of the old model. Most existing exemplar-free methods are based on parameter regularization, which leads to unsatisfactory performance. On the other hand, current rehearsal-based methods achieve good performance by accessing data of previous tasks explicitly or implicitly, which is often not allowed in real world applications. We are committed to address the dilemma by synthesizing informative pseudo data in a strictly data-free manner for incremental learning.

\subsection{Self-distilled Knowledge Delegator}
\begin{figure}[t]
\centering
\includegraphics[width=0.8\columnwidth]{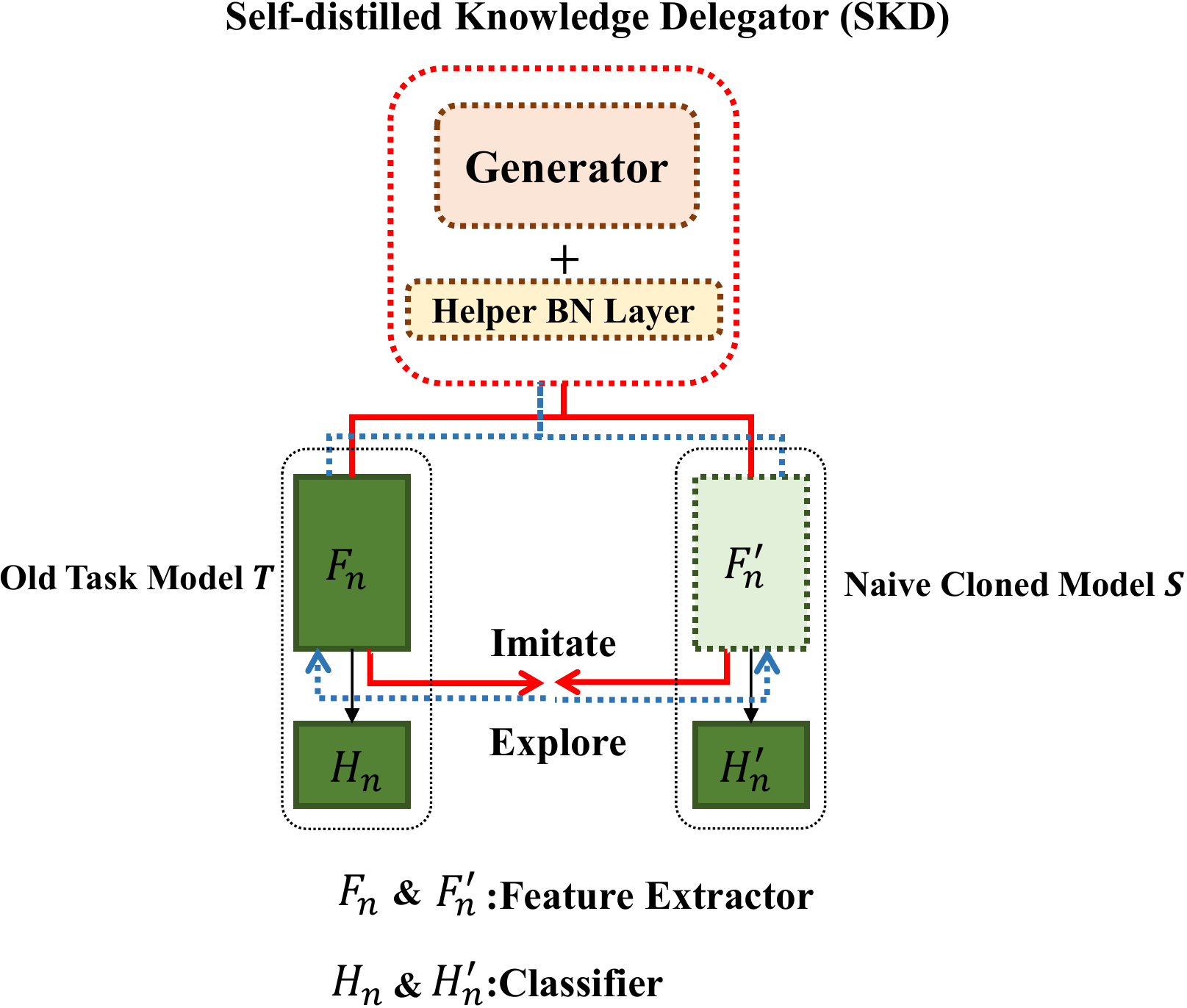}
  \caption{The imitate \& explore adversarial training process of SKD. SKD acts as a knowledge delegator to transfer the knowledge from ${T}$ to ${S}$. After training, $S$ is dropped, and SKD is adopted for incremental learning.}
\label{trainSKD}
\end{figure}

In order to maintain the performance of the model on old tasks as much as possible while being plastic on new data, the knowledge encoded in the model trained over old tasks should be extracted. We propose the Self-distilled Knowledge Delegator (SKD) to do this consolidation by synthesizing informative pseudo data. As shown in Fig.~\ref{trainSKD}, SKD is essentially a generator followed by a helper BN layer. It is trained using the self-distillation mechanism. That is, given a pre-trained model ${T_n}$ on task ${n}$, we attempt to transfer its knowledge to a naive cloned model ${S_n}$ with the help of SKD. ${S_n}$ shares the same architecture as ${T_n}$, but its parameters are re-initialized randomly. Based on the informative data generated by SKD, ${S_n}$ is expected to retain the feature extraction capability as ${T_n}$. To accomplish this, a novel imitate \& explore adversarial training strategy is introduced. Notably, SKD is distinct from the generator of classic GAN~\cite{mirza2014conditional} in the following aspects. 1) The training process of SKD is completely data-free. 2) SKD aims to generate pseudo data that help a model to recover its feature extraction capability rather than obey the distribution of real samples. These properties make it perfectly fit the problem of exemplar-free CIL, where the key lies in overcoming the forgetting of classification capability on old classes.

\subsubsection{Imitate \& Explore Adversarial Training Strategy}
The imitate \& explore adversarial training strategy consists of two phases as its name suggests. The imitate phase minimizes the discrepancy between ${T_n}$ and ${S_n}$, while the explore phase does the opposite by forcing SKD to generate more informative data. During the training process, the weights of the entire pre-trained old model ${T_n}$ are fixed. SKD is only updated in the explore phase, and ${F_n^{'}}$ is kept updated throughout the whole imitate and explore phases.

\textbf{Feature Cosine Discrepancy} The vanilla knowledge transfer methods such as~\cite{hinton2015distilling} define the discrepancy on the output logits by measuring the Kullback-Leibler divergence. On the contrary, we define the feature discrepancy between two models on a normalized feature hyper-sphere, named \emph{feature cosine discrepancy} by:
\begin{equation}\label{cos_discrepancy}
  {\psi}\left( {\bf{x}'}, {T}_n,{S}_n \right) = 1-\cos \left( {nor\left( {{{{F}_n}}\left( \bf{x}{'} \right) } \right),nor\left( {{{{F}_n{'}}}\left( \bf{x}{'} \right) } \right)} \right),
\end{equation}
where ${{{F}_n}\left( \cdot \right)}$ and ${{{F}_n{'}}\left( \cdot \right)}$ are the feature extractors of ${{T}_n}$ and ${{S}_n}$ respectively. $nor\left(  \cdot \right)$ denotes L2 normalization, and  $\cos \left(  \cdot  \right)$ is the cosine similarity function. ${\bf{x}{'}}$ is the pseudo data generated by SKD.

\textbf{Imitate Phase:} In the imitate phase, the naive cloned model ${{{S}}_n}$ tries to mimic ${{{T}_n}}$ by minimizing the feature cosine discrepancy (denoted as ${{{\cal L}_{imi}}}$ in Eq.~\eqref{loss_imi}) given any output of SKD. Intuitively, arbitrary data generated by a weak SKD model cannot guarantee ${{{S}}_n}$ to be equipped with the similar feature extraction capability as ${{{T}_n}}$, because the pseudo data are too uncontrollable to fall within the feature distribution of real data. Once SKD could produce enough data that match with the real data in terms of feature distribution in latent space, such superior SKD can be considered to be able to generate informative data.
\begin{equation}\label{loss_imi}
{{\cal L}_{imi}} =  {\psi}\left({\bf{x}'},{{{T}_n},{{S}_n}} \right)
\end{equation}

\begin{figure}[t]
\centering
\includegraphics[width=0.45\columnwidth]{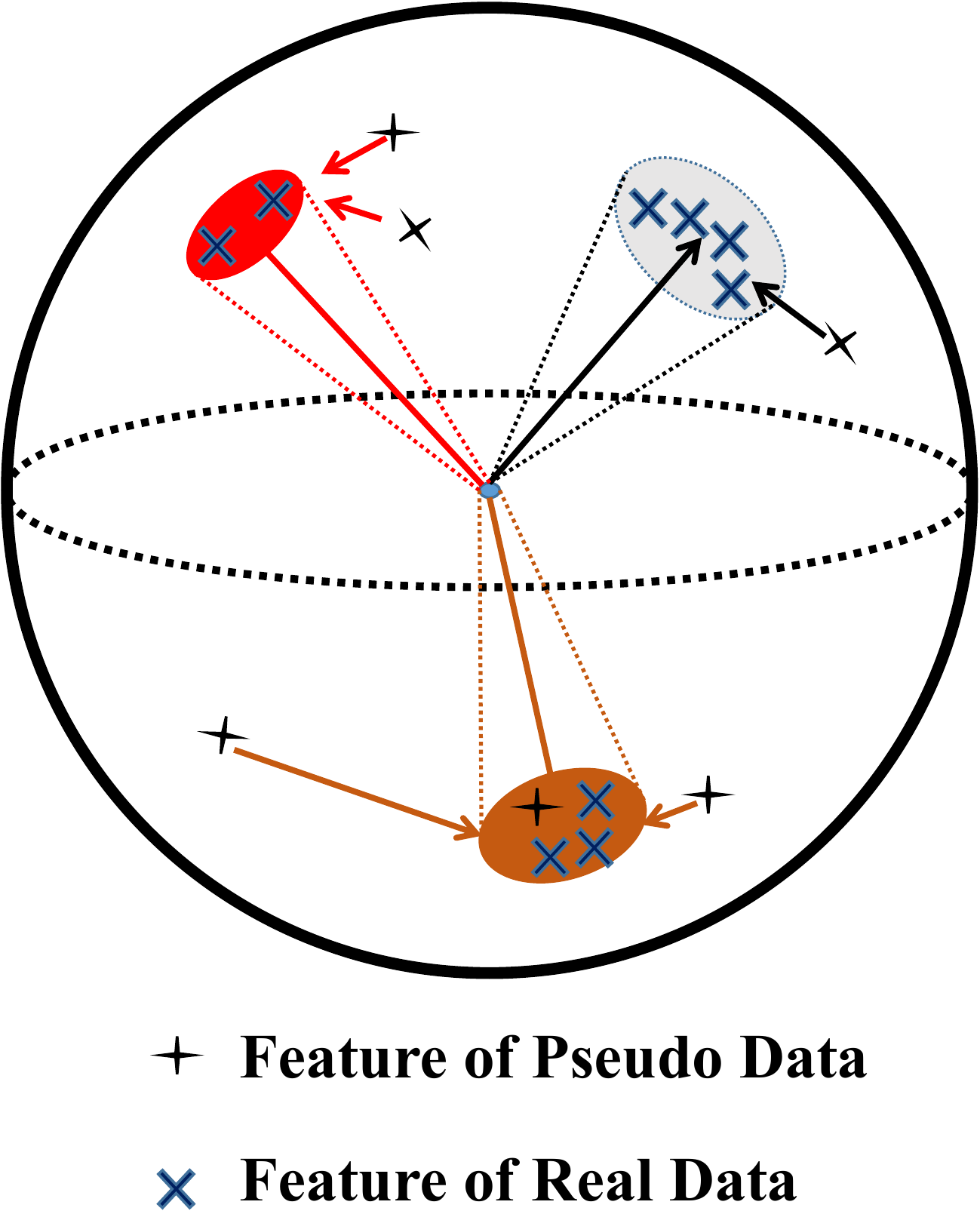}
  \caption{The distribution of features of real data and pseudo data in the latent hyper-spherical space. $\theta$ is the angle between two classes. The arrows represent the moving direction of the pseudo data.}
\label{featCircle}
\end{figure}

\textbf{Explore Phase:} In the explore phase, SKD is encouraged to produce more informative data, so that the features of pseudo data are well aligned with those of real data. Given a pre-trained model, the feature distribution of real data in the latent space has an obvious clustering effect. As demonstrated in Fig.~\ref{featCircle}, features of the same class will be compactly aggregated in a small area as indicated by the shaded regions on the hyper-sphere. However, the features of data produced by a weak SKD model are likely to be randomly scattered in the latent space (marked as black stars in Fig.~\ref{featCircle}). The explore phase is marked as arrows in Fig.~\ref{featCircle}, which essentially pushes the features of synthetic data towards features of real data.

In the explore phase, in order to drive SKD to explore more informative data, an exploration loss ${{\cal L}_{exp}}$ is first proposed to push ${{S}_n}$ away from ${{T}_n}$ by enlarging the feature cosine discrepancy.

\begin{equation}\label{loss_exp}
{{\cal L}_{exp}} =  -{{\cal L}_{imi}}
\end{equation}
Eq.~\eqref{loss_exp} optimizes towards the opposite direction of Eq.~\eqref{loss_imi}. SKD can be regarded as a data explorer that traverses the data space where real data only occupy a small area. Notice that under the restriction of inaccessible to any real data, our explore phase should expand the scope of SKD's exploration to cover the distribution of real data. To help SKD to generate informative data that match the feature distribution of real data in latent space, three auxiliary objectives are introduced.

To make SKD generate category-aware samples rather than an average image of all classes, we adopt the cross-entropy loss to guarantee that data generated by SKD can be classified into an arbitrary category with high probability.
\begin{equation}\label{loss_cat}
{{\cal L}_{cat}} = {E_{z \sim p\left( z \right)}}\left( { - \sum\limits_k^K {{{y'}_k}\log \left( {H_n^k\left( {{F_n}\left( {D\left( z \right)} \right)} \right)} \right)} } \right),
\end{equation}
where ${z}$ denotes the randomly latent vector sampled from the Gaussian distribution ${p\left( z \right)}$. $H_n^k$ is the score of class $k$ predicted by $T_n$. ${\bf{x}'}=D(z)$ is the pseudo data generated by SKD. For each pseudo datum, the pre-trained model ${{{T}_n}}$ is used to label it, which is formulated as:
\begin{equation}\label{loss_onehot}
y' = \emph{one-hot}\left( {\arg \max \left( T_n({\bf{x}'}) \right)} \right),
\end{equation}
where $y'$ is a one-hot vector that denotes the pseudo label of each pseudo sample. ${{\cal L}_{cat}}$ reduces the risk of random exploration, and helps SKD to focus on samples relevant to real data in the label space.

Besides, to avoid the occurrence of mode collapse, we must ensure the diversity of pseudo data by SKD from the perspective of ${{T}_n}$. So another \emph{diversity loss} is proposed as:
\begin{equation}\label{loss_div}
\begin{split}
{{\cal L}_{div}} &= \sum\limits_k^K {{w_k}\log \left( {{w_k}} \right)}\\
{w_k} &= E_{z \sim p\left( z \right)}\left( \frac{1}{B}{\sum {H_n^k({F_n}\left( {D(z)} \right))}}\right),
\end{split}
\end{equation}
where $B$ denotes the batch size, and the second summation is performed over the classification score of current batch. ${{\cal L}_{div}}$ encourages the data in a batch to be uniformly distributed across different classes.

Inspired by \cite{Yin2020}, the \emph{feature distribution regularization term} is introduced to guide SKD to generate data with similar distribution to the real training data. As we know, the BN layers of a pre-trained model contain statistics of the running mean and variance of the training data. The feature distribution regularization term is denoted as:
\begin{equation}\label{loss_rfeat}
\begin{split}
{{\cal R}_{feature}} = & {E_{z \sim p\left( z \right)}} \left( {\sum\limits_l {\left\| {{\mu _l}\left( {D(z)} \right) - \hat{\mu}_l} \right\|} _2}\right) +\\
 &{E_{z \sim p\left( z \right)}} \left( {\sum\limits_l {\left\| {{\sigma _l}\left( {D(z)} \right) - \hat{\sigma}_l} \right\|} _2}\right)
\end{split}
\end{equation}
where $\hat{\mu}_l$ and $\hat{\sigma}_l$ are the running mean and variance of the $l$-th Batch-Normalization layer (BN layer) in the pre-trained $T_n$, and ${\mu _l}\left( {D(z)} \right)$ and ${\sigma _l}\left( {D(z)} \right)$ are the running mean and variance in $S_n$ to be trained.

The total loss in the explore phase is given as:
\begin{equation}\label{loss_skd_all}
{{\cal L}_{explore}} = \lambda {{\cal L}_{exp}}{\rm{ + }}{{\cal L}_{cat}} + {{\cal L}_{div}} + {{\cal R}_{feature}},
\end{equation}
where ${\lambda}$ is a parameter set according to different adversarial space angles ${\theta}$ empirically as illustrated in Fig.~\ref{featCircle}. ${\theta}$ represents the inter-class distance, which varies on different datasets according to our experimental studies. Intuitively, the more fine-grained the dataset is, the smaller ${\theta}$ will be. A smaller ${\theta}$ requires a larger ${\lambda}$ to balance different constraints.

\subsubsection{The Helper BN Layer}
As shown in Fig.~\ref{trainSKD}, we append a helper BN layer after the generator in SKD. The helper BN layer is critical since there might be a large domain drift from real data for the output of the generator under data-free training. This is harmful for both training of SKD and the subsequent CIL. The helper BN layer can help to normalize the distribution of the output of the generator to approximate the distribution of real data (after preprocessing). Hence, the domain conflict between pseudo data and real data will be greatly alleviated.

\subsubsection{Necessity of the Constraints}

The constraints introduced above are critical to the success of SKD considering the training difficulty without real data. To shape the generated pseudo data by SKD, we apply abundant regularizations at various levels. The helper BN layer regulates the output data of SKD directly. ${{\cal R}_{feature}}$ constrains SKD from the perspective of feature statistics in the intermediate layers. And ${{\cal L}_{cat}}$ and ${{\cal L}_{div}}$ constrain the final probability output of $T_n$. As discussed in Section~\ref{sect:abl-stu}, these techniques are all important for the training of SKD.

\subsection{SKD-based CIL}

Given the SKD model trained by the proposed imitate \& explore adversarial training strategy, we are able to consolidate the model performance on tasks already learned. According to our experiments, a naive cloned model ${{S}_n}$ can be taught to achieve almost the same feature extraction capabilities as ${{T}_n}$. Based on the strong knowledge transfer capability of SKD, we design a CIL method. The training procedure is similar to exemplar-based methods, where the exemplars are replaced by the pseudo data generated by SKD. In this section, we drop the notation $D(z)$ for simplicity since SKD is fixed and used to generate pseudo data.

Inspired by~\cite{Hou2019, Li2018}, we adopt the feature consolidation loss and cross entropy loss to allow the model to learn new knowledge while preventing catastrophic forgetting. Before training incrementally on task ${n+1}$, ${{T}_{n+1}}$ is initialized from ${{T}_{n}}$ by copying their weights. A data batch consists of both real data from new task and pseudo data from SKD, \ie ${\bf{\hat x}} = \{{\bf{x}}',{{\bf{x}}_{{new}}}\}$. To alleviate the data imbalance issue, the numbers of pseudo and real data are kept the same in a single batch.

Based on feature cosine discrepancy, the feature consolidation loss aims to maintain the performance on previous tasks using both new data and pseudo data produced by SKD, which is given as
\begin{equation}\label{loss_fc}
\begin{split}
  {{\cal L}_{fc}} &= {\psi}\left( {\bf{\hat x}},{{{T}_n},{{ T}_{n+1}}} \right) \\
  &= 1-\cos \left( {nor\left( {{{{F}_n}}\left( \bf{\hat x} \right) } \right),nor\left( {{{{F}_{n+1}}}\left( \bf{\hat x} \right) } \right)} \right).
\end{split}
\end{equation}
And the cross-entropy classification loss is formulated as
\begin{equation}\label{loss_cls}
{{\cal L}_{cls}} = - \sum\limits_k^K {{{\hat y}_k}\log \left( {{{H}_{n + 1}^k}\left( {{{\hat x}_k}} \right)} \right)}.
\end{equation}
Notice that the labels of pseudo data are given by ${T}_n$ as in Eq.~\eqref{loss_onehot}.
The total loss in each CIL task is given as
\begin{equation}\label{loss_cil}
{{\cal L}_{cil}} = \gamma {{\cal L}_{cls}}{\rm{ + }}{{\cal L}_{fc}}.
\end{equation}

Obviously, for different datasets in different incremental tasks, ${{\cal L}_{cls}}$ and ${{\cal L}_{fc}}$ need to be given different weights. In order to balance stability and plasticity, we adopt a simple yet effective adaptive loss weighting schedule defined as
\begin{equation}\label{loss_gama}
  {\gamma}_n  = \frac{\beta }{N\cdot{\sum_{i=1}^n|{{\cal C}_{i}}|}},
\end{equation}
where $n$ and ${N}$ denote the current task id and total number of tasks respectively. $|\cdot|$ is the cardinality operator and $\beta$ is a fixed constant for each dataset. $\gamma $ decays as the number of accumulated new categories increases.

Notably, although the training procedure of CIL here is similar to previous works~\cite{Hou2019}, there are two significant differences. 1) The proposed method is completely exemplar-free and does not need to store and access any data of previous tasks. 2) We do not require any special parameter initialization method and metric learning strategy, where the \emph{imprint weights} and \emph{inter-class separation} play an extremely important role in~\cite{Hou2019}.

\subsection{Implementation}

\begin{algorithm}[t]
\caption{Self-distilled Knowledge Delegator for Exemplar-free Class Incremental Learning}\label{algorithm1}
\KwIn{A pre-trained model ${T}_0$.}
\KwOut{Model ${T}_{n+1}$ after $n+1$ incremental tasks.}
\For{$n\leftarrow 0$ \KwTo $N-1$}{
\textbf{1. Train SKD:}\\
\eIf{$n > 0$}{
      Initialize $D_{n}$ with $D_{n-1}$;\\
      }{
      Initialize $D_{0}$ randomly;
      }
\For{number of SKD training epochs}{
\For{$5$ $steps$}{Fix ${D}_{n}$, train ${{S}_n}$ by minimizing ${{\cal L}_{imi}}$ in Eq.~\eqref{loss_imi};}
Train both ${{S}_n}$ and ${D}_{n}$ by minimizing ${{\cal L}_{explore}}$ in Eq.~\eqref{loss_skd_all};
}
\textbf{2. Train CIL:}\\
Initialize ${{T}_{n+1}}$ with ${{T}_{n}}$;\\
\For{number of CIL epochs}{
Train ${{T}_{n+1}}$ by minimizing ${{\cal L}_{cil}}$ in Eq.~\eqref{loss_cil};
}
}
\end{algorithm}

We summarize the training of the proposed incremental learning framework in Algorithm~\ref{algorithm1}. For each CIL task, the input is just a model trained on the previous task. The algorithm consists of two stages, namely training of SKD and training of the incremental model. When training SKD, the number of training epochs in the imitate phase and the explore phase are different. A single SKD model is continually fine-tuned over sequential tasks.

\section{Experiments}
We conduct extensive experiments and ablation studies to verify the effectiveness of our method on four datasets. %\ie ImageNet-Subset~\cite{Rebuffi2017}, Caltech-101~\cite{fei2004learning} and Flowers-102~\cite{nilsback2008automated}.

\subsection{Datasets and Implementation Details}
\textbf{Datasets:} ImageNet-Subset, following~\cite{Hou2019, Liu2020}, is built by randomly sampling 100 classes from ImageNet with a random seed (1993) using the NumPy library~\cite{Harris2020aa}. CIFAR-100 contains 100 classes with 600 images per class. There are 8189 images in Flowers-102, and 6982 images in Caltech-101~\cite{fei2004learning}. The images are resized to $224 \times 224 $ for ImageNet-Subset, $32 \times 32 $ for CIFAR-100 and $128 \times 128 $ for both Flowers-102 and Caltech-101. The category order of all datasets will be randomly permuted by a specific seed, \ie 1993 for CIFAR-100 and 1 for both Flowers-102 and Caltech-101. For all datasets, we start from a model pre-trained on 50 classes as used in~\cite{Hou2019, Liu2020}.

\textbf{Implementation Details:} All models and experiments are implemented using PyTorch~\cite{paszke2017automatic}. SGD~\cite{bottou2010large} is used in the imitate phase of SKD, and the initial learning rate is set to 0.1. In the explore phase, Adam~\cite{Addepalli2019} is used and the initial learning rate is set to 0.001. The learning rate drops by 10 times every 100 epochs. When training SKD, the batch size of pseudo data is set to 256 for CIFAR-100 and Caltech-101, and 1024 for other datasets. When training the incremental model, the SGD optimizer is adopted, and the learning rate starts from 0.1 and is divided by 10 after 80 and 120 epochs (160 epochs in total). We adopt a 32-layer ResNet for CIFAR-100 and an 18-layer ResNet for other datasets. For other hyper-parameters, $\lambda$ is set to 20.0 for the fine-grained dataset Flowers-102 and 1.0 for the rest datasets, and ${\beta}$ is set to 5.0 for all datasets. The backbone of SKD is lightweight and kept the same as \cite{Fang2019}.

\subsection{Ablation Studies}
\label{sect:abl-stu}

\begin{figure}[t]
\centering
\subfigure[Real data]{
	\begin{minipage}[t]{0.5\linewidth}
		\centering
		\includegraphics[width=1.6in]{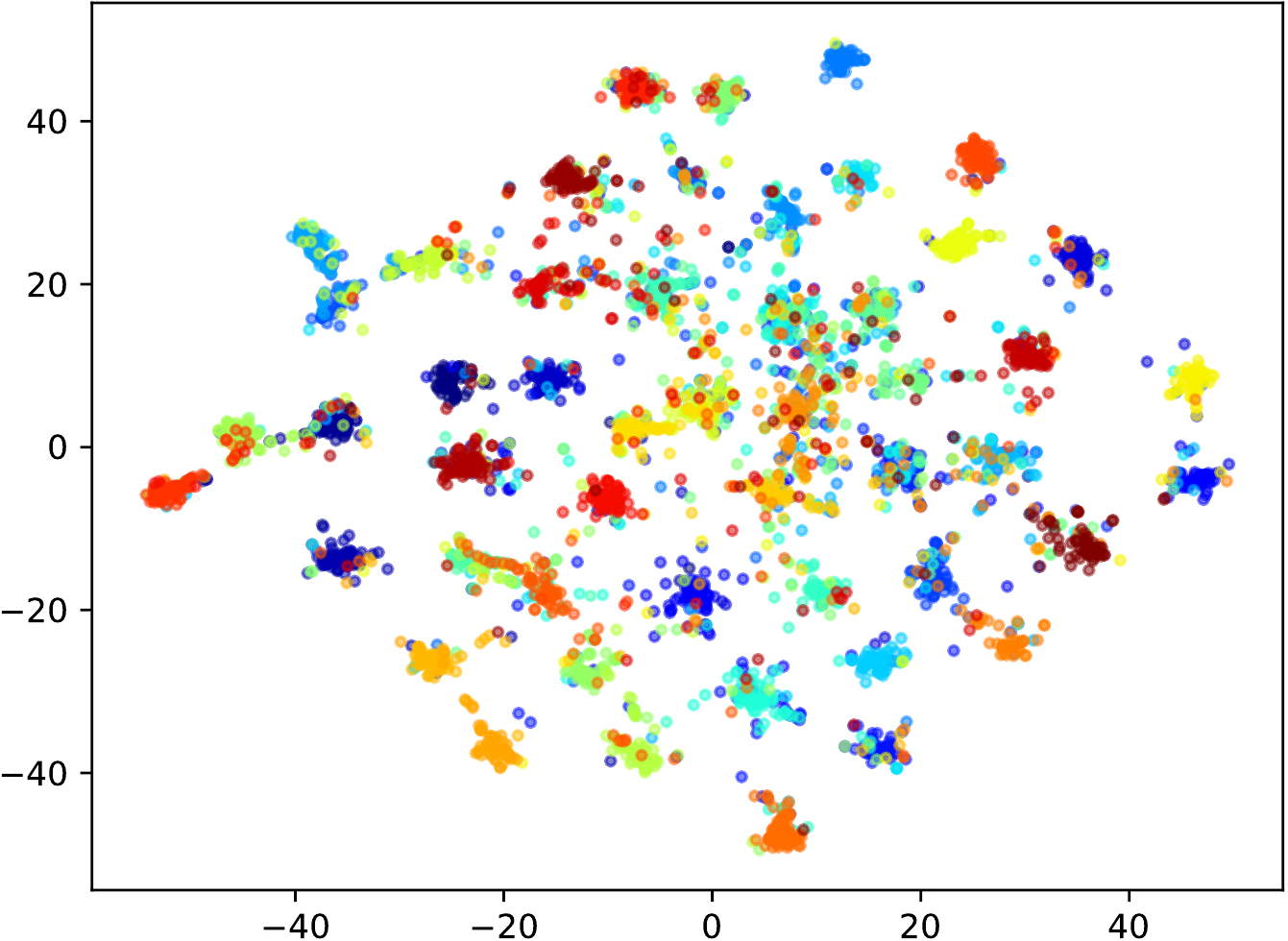}
		%\caption{fig1}
	\end{minipage}%
	\label{tsne_a}
}%
\subfigure[Pseudo samples by SKD]{
	\begin{minipage}[t]{0.5\linewidth}
		\centering
		\includegraphics[width=1.6in]{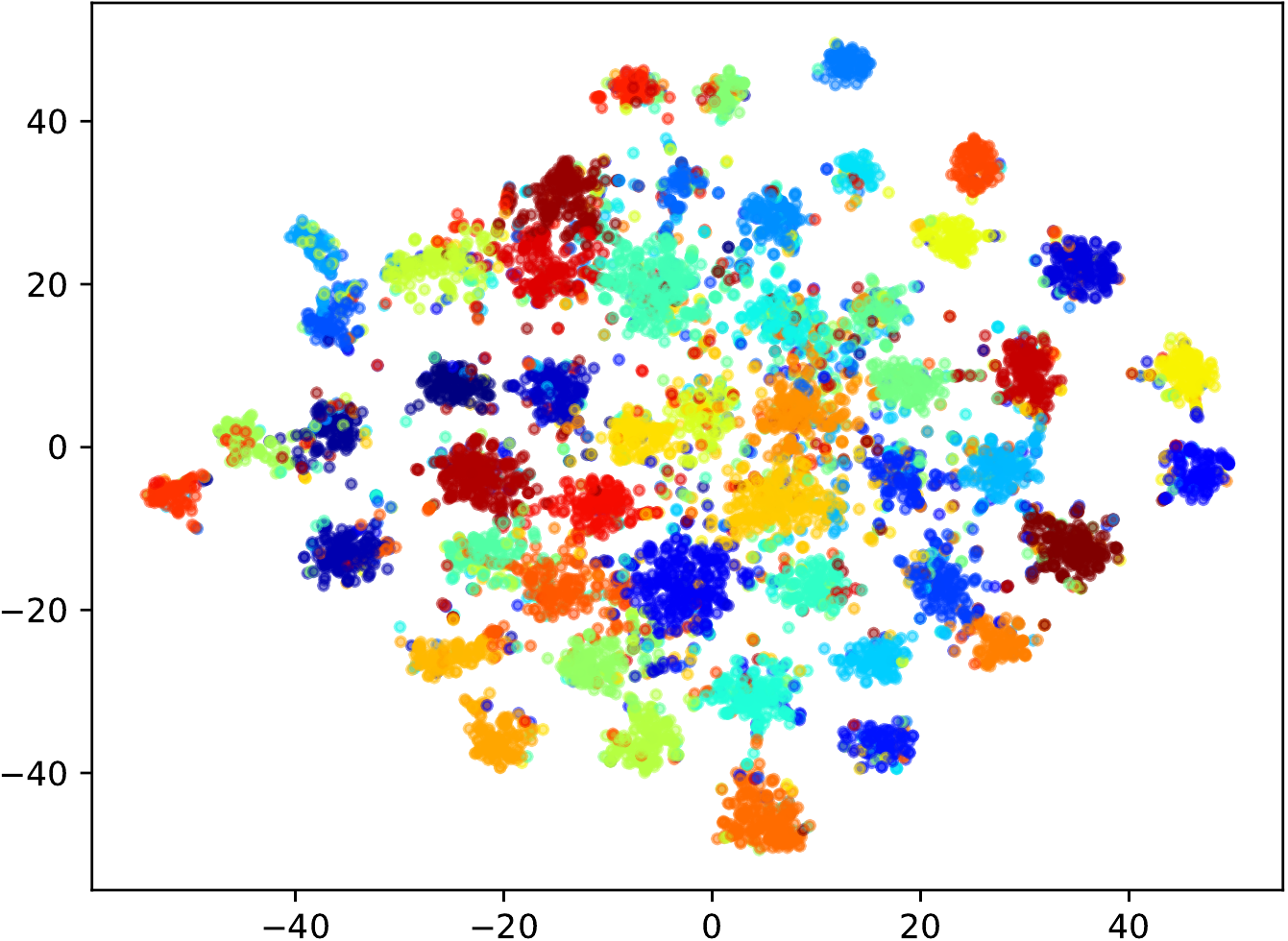}
		%\caption{fig2}
	\end{minipage}%
	\label{tsne_b}
}%

\subfigure[Samples by SKD w/o ${{\cal L}_{cat}}$]{
	\begin{minipage}[t]{0.5\linewidth}
		\centering
		\includegraphics[width=1.6in]{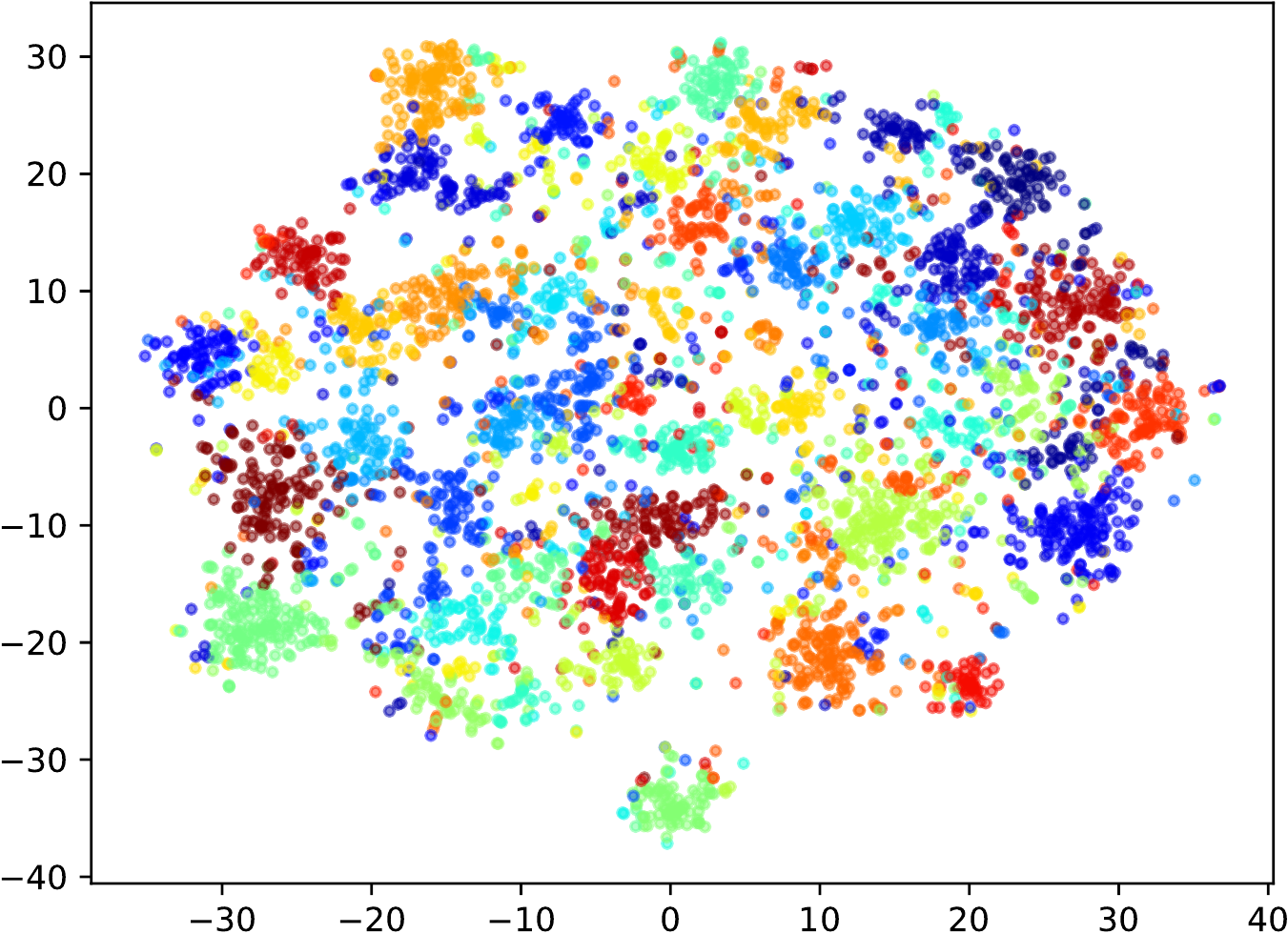}
		%\caption{fig1}
	\end{minipage}%
	\label{tsne_c}
}%
\subfigure[Samples by SKD w/o ${{\cal L}_{div}}$]{
	\begin{minipage}[t]{0.5\linewidth}
		\centering
		\includegraphics[width=1.6in]{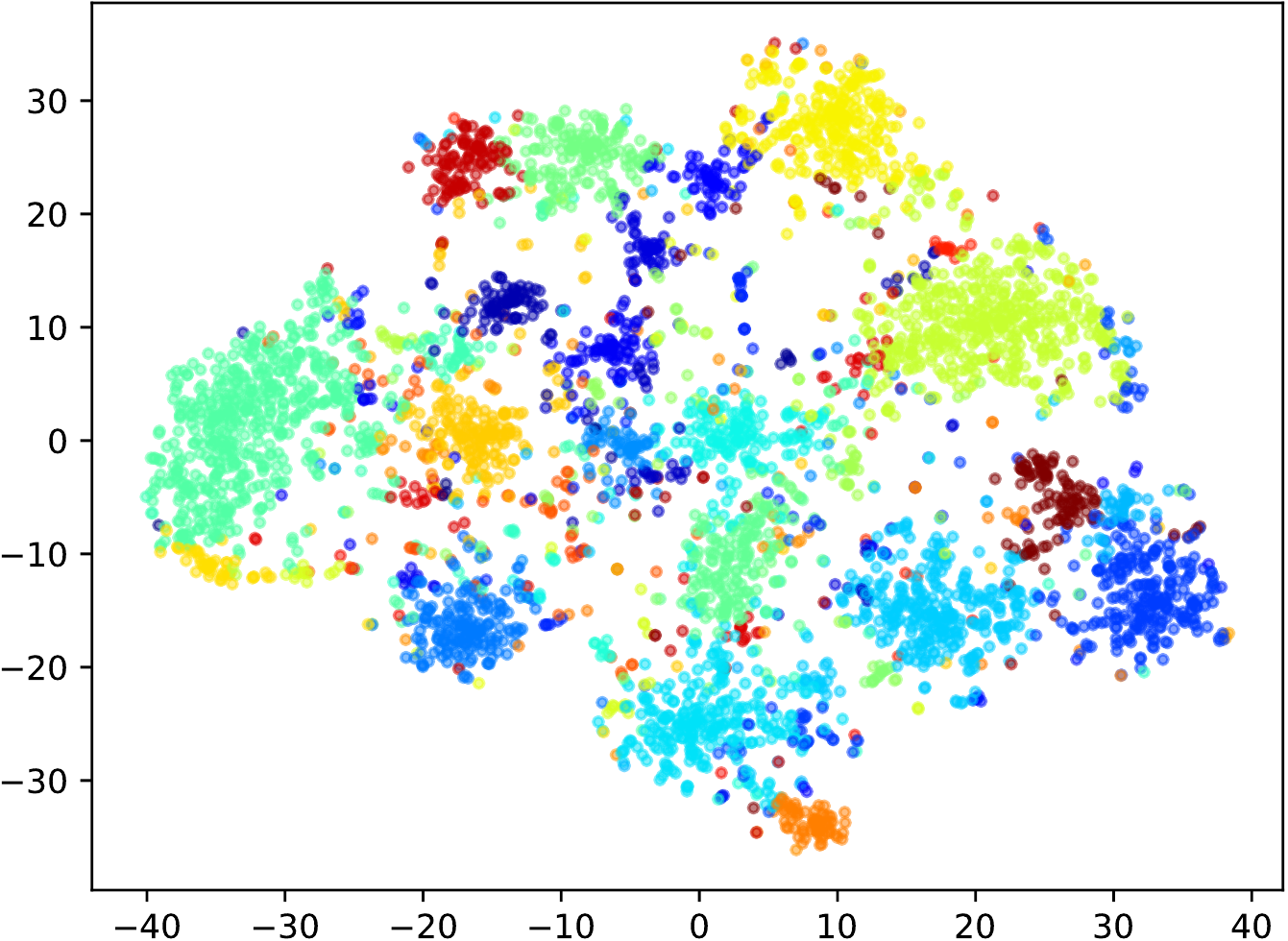}
		%\caption{fig2}
	\end{minipage}%
	\label{tsne_d}
}%
\centering
\caption{t-SNE visualization of real data and pseudo samples generated by SKD in the feature space from the perspective of ${T}_0$. Different classes are labeled in different colors.}
\label{tsne}
\end{figure}

\textbf{Knowledge Transfer Capability of SKD.}
As discussed earlier, the knowledge transfer capability of SKD from the pre-trained model $T_n$ to the cloned model $S_n$ is the fundamental premise for the success of the subsequent incremental learning. To validate this capability, ex post facto recognition experiments and visualizations are conducted. We use ${{H}_0}$ of ${{T}_0}$ to initialize classifier ${{H}_0{'}}$ of ${{S}_0}$, then measure the accuracy of $S_0$ on the four datasets and compare it with $T_0$. As shown in Table~\ref{accgap}, the proposed SKD is able to reduce the accuracy gap between $T_0$ and $S_0$ to within 1\%. Notably, in the following CIL tasks, this knowledge transfer capability can be well maintained on most datasets. Remarkably, SKD is very robust against the scale of the problem in terms of input resolution (from $32\times 32$ to $224\times 224$) and network architecture, which is also a significant advancement over existing knowledge transfer works~\cite{Chen2019, Choi2020, Fang2019, Micaelli2019, Nagel2019}. As visualized in Fig.~\ref{tsne_a} and~\ref{tsne_b}, from the perspective of $T_0$ the feature distribution of real data and pseudo data are almost the same.

\begin{table}[t]
  \centering
  \caption{The accuracy of $T_0$ and ${S}_{0}$ on four different datasets.}
  \scalebox{1.0}{
    \begin{tabular}{c|c|c|c}
    \hline
    Datasets & ${T}_{0}$ & ${S}_{0}$  & Acc Gap \\
    \hline
    CIFAR-100 & 76.08\%  & 75.86\%  & 0.22\% \\
    \hline
    Caltech-101 & 76.64\%  & 75.90\%  & 0.74\% \\
    \hline
    Flowers-102 & 84.54\%  & 83.91\%  & 0.63\% \\
    \hline
    ImageNet-Subset & 87.68\%  & 86.80\%  & 0.88\% \\
    \hline
    \end{tabular}%
  \label{accgap}}%
\end{table}%

\textbf{Impact of the Training Techniques for SKD.} In the imitate \& explore adversarial training strategy, the explore phase is critical for SKD to learn high quality pseudo data. To show the importance of the auxiliary losses (${{\cal L}_{cat}}$ and ${{\cal L}_{div}}$), the regularization term (${{\cal R}_{feature}}$) and the helper BN layer, we measure the accuracy of $S_n$ when each of them is disabled. As shown in Table~\ref{ablation_skd}, these techniques barely affect the training of SKD on the relatively simple CIFAR-100 dataset. However, on more complex datasets such as ImageNet-Subset, they are critical for the success of SKD. The training of SKD totally fails without ${{\cal R}_{feature}}$. Removing ${{\cal L}_{cat}}$ leads to unstable training and significant performance degradation. ${{\cal L}_{div}}$ and the helper BN affect both the convergence speed and the final performance.

\begin{table}[t]
  \centering
  \caption{Impact of the four training techniques to accuracy of $S_0$ on CIFAR-100 and ImageNet-Subset.}
  \scalebox{1.0}{
    \begin{tabular}{c|c|c}
    \hline
          Settings & CIFAR-100 & ImageNet-Subset \\
    \hline
    Upper Bound Acc & 76.08\%  & 87.68\%  \\
    \hline
    SKD w/o ${{\cal L}_{cat}}$ & 75.76\%  & 46.72\%  \\
    \hline
    SKD w/o ${{\cal L}_{div}}$ & 76.00\%  & 82.68\%  \\
    \hline
    SKD w/o ${{\cal R}_{feature}}$ & 75.34\%  & 2.84\% \\
    \hline
    SKD w/o helper BN & 76.01\%  & 85.68\%  \\
    \hline
    SKD (Ours) & 75.86\%  & 86.80\%  \\
    \hline
    \end{tabular}%
  \label{ablation_skd}}%
\end{table}%

To verify our assumption on strong SKD, we adopt ${{T}_0}$ as an example to extract features of both pseudo data generated by SKD and real training data on CIFAR-100, and then compare their distribution in the feature space with t-SNE~\cite{maaten2008visualizing}. As shown in Fig.~\ref{tsne}, obviously the category orientation and diversity of the data generated by SKD are significantly degenerated without ${{\cal L}_{cat}}$ and ${{\cal L}_{div}}$. Equipped with all of the techniques, SKD is capable of generating pseudo data that exhibit consistent feature distribution to the original training data. Note that we aim to guarantee similar distribution in the feature space from the perspective $T_n$. 

\begin{table}[t]
  \centering
  \caption{Impact of SKD and ALW to average accuracies of 5 tasks and 10 tasks on CIFAR-100.}
  \scalebox{1.0}{
    \begin{tabular}{c|c|c}
    \hline
    Settings & 5 Tasks & 10 Tasks \\
    \hline
    w/o SKD \& ALW & 29.16\%  & 18.26\%  \\
    \hline
    w/o SKD & 56.35\%  & 49.96\%  \\
    \hline
    Ours  & 61.42\%  & 59.62\%  \\
    \hline
    \end{tabular}%
  \label{ablation}}%
\end{table}%

\textbf{Importance of SKD for CIL.} The goal of SKD is to consolidate the performance of the model on old tasks, thereby alleviating catastrophic forgetting. As shown in Table~\ref{ablation}, when the SKD module is disabled, the incremental learning performance of the model degrades greatly on CIFAR-100 in both 5-task and 10-task settings. 

\textbf{Importance of Adaptive Loss Weighting.} The proposed adaptive loss weighting schedule (Eq.~\eqref{loss_gama}) can automatically balance the feature consolidation loss and cross-entropy loss for different datasets in different tasks. This trick is critical to achieve a reasonable baseline performance. As shown in Table~\ref{ablation}, when it is disabled, the loss weighting basically degenerates into the way in LwF~\cite{Li2018}, and the performance drops significantly.

\subsection{Comparison with the State-of-the-arts}
\begin{figure*}[]
\centering
\subfigure[CIFAR-100 (1 task)]{
	\begin{minipage}[t]{0.24\linewidth}
		\centering
		\includegraphics[width=1.6in]{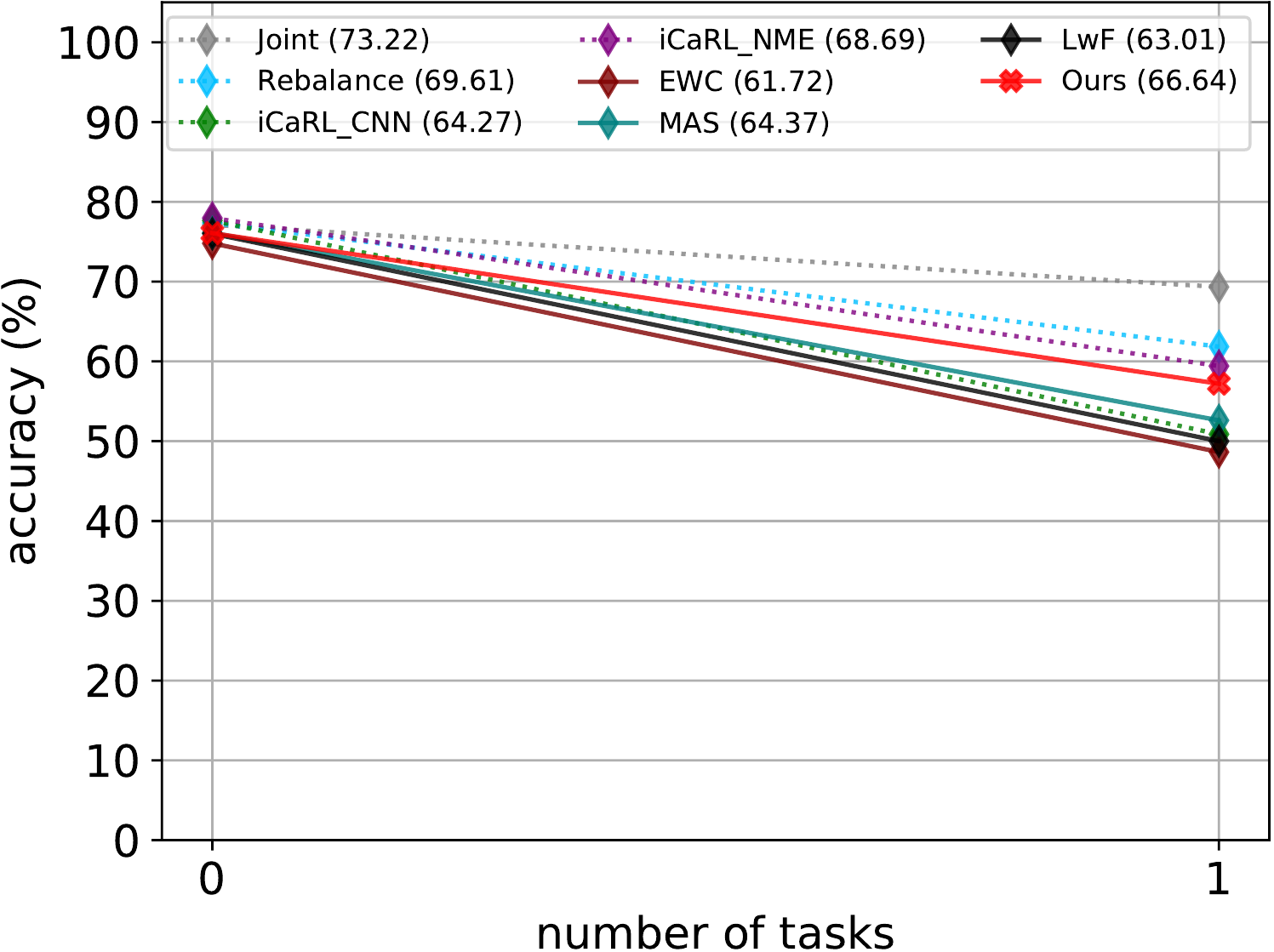}
		%\caption{fig1}
	\end{minipage}%
	\label{cifar_1}
}%
\subfigure[CIFAR-100 (2 tasks)]{
	\begin{minipage}[t]{0.24\linewidth}
		\centering
		\includegraphics[width=1.6in]{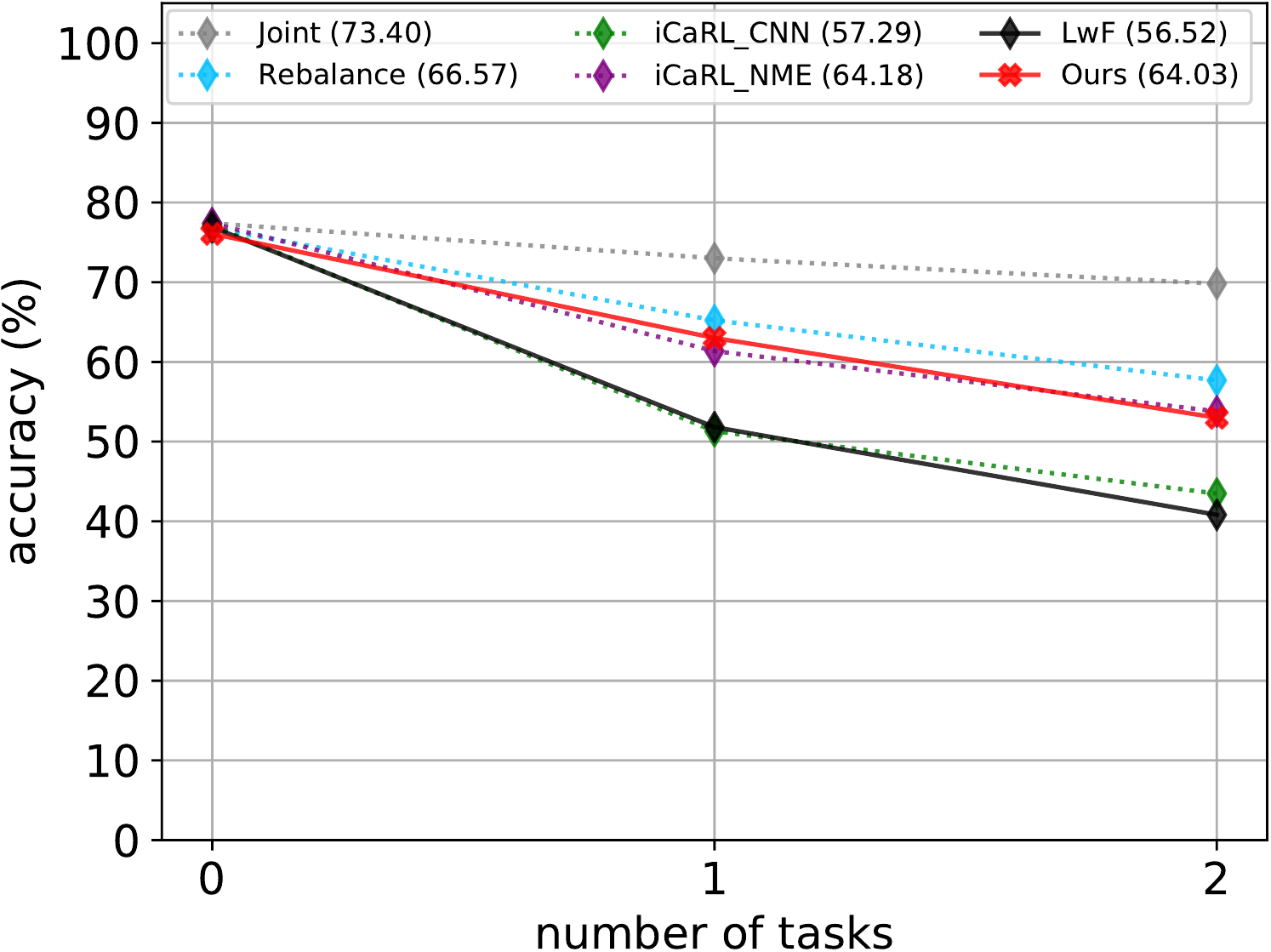}
		%\caption{fig2}
	\end{minipage}%
	\label{cifar_2}
}%
\subfigure[CIFAR-100 (5 tasks)]{
	\begin{minipage}[t]{0.24\linewidth}
		\centering
		\includegraphics[width=1.6in]{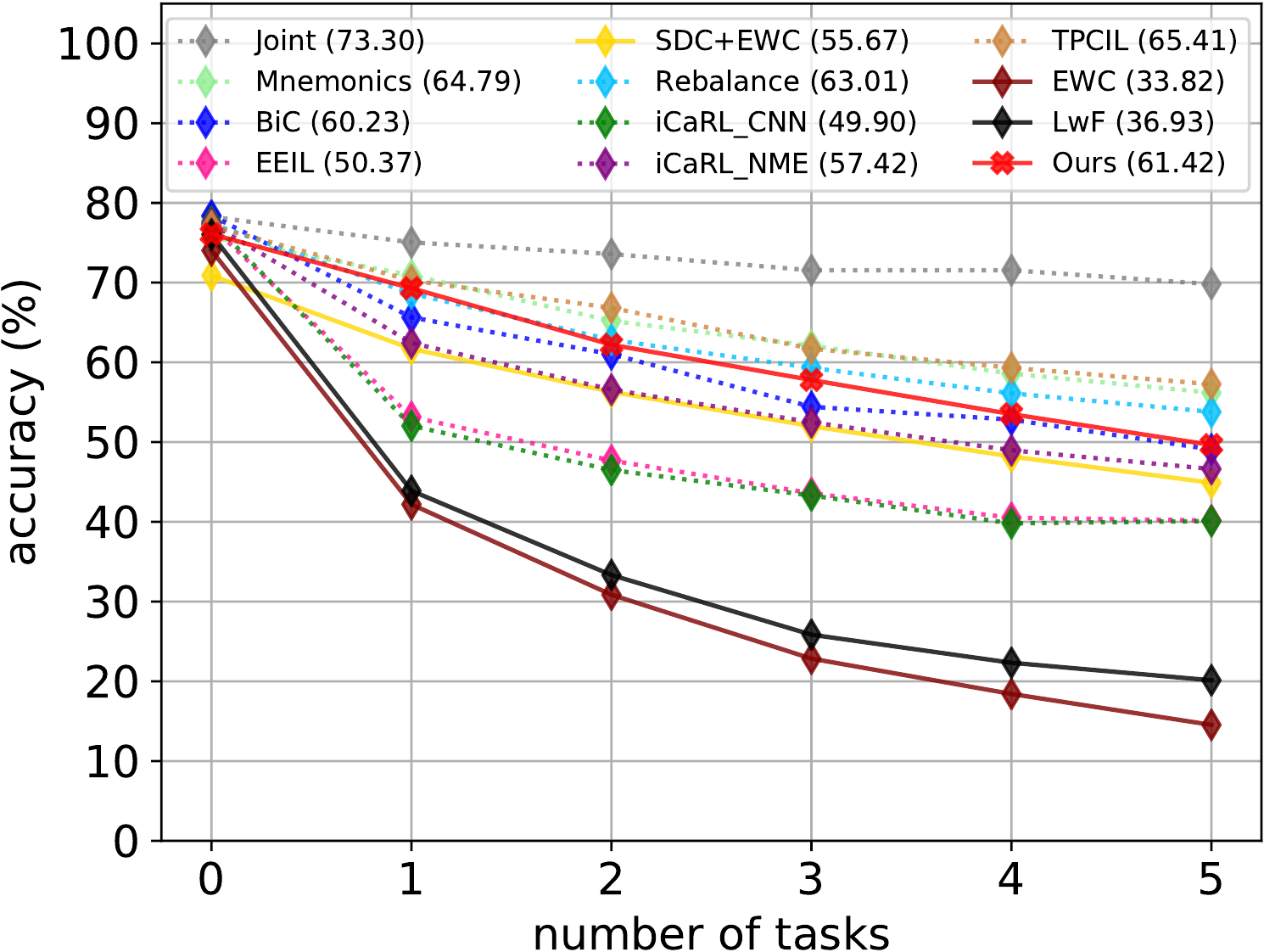}
		%\caption{fig2}
	\end{minipage}
	\label{cifar_5}
}%
\subfigure[CIFAR-100 (10 tasks)]{
	\begin{minipage}[t]{0.24\linewidth}
		\centering
		\includegraphics[width=1.6in]{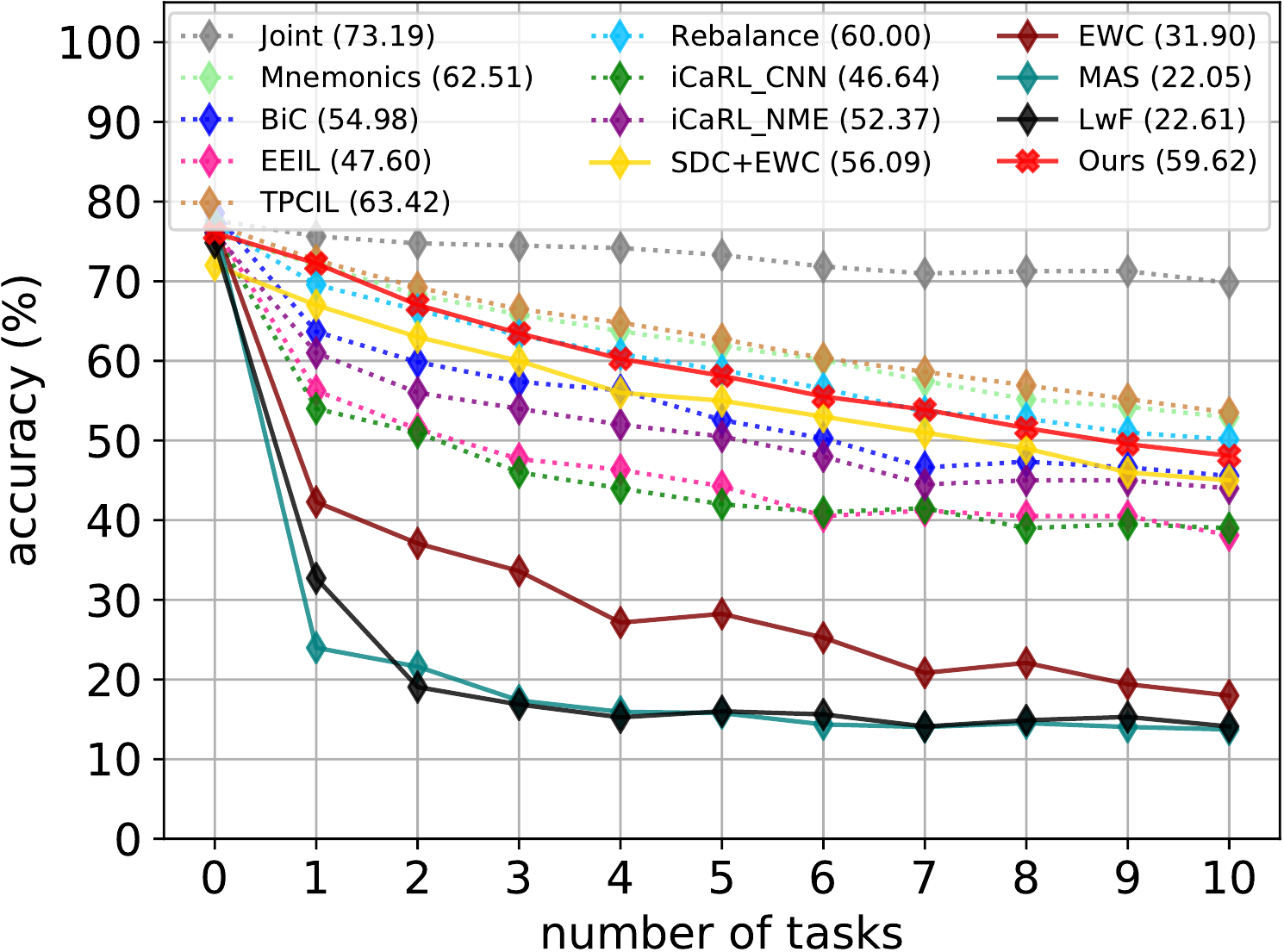}
		%\caption{fig2}
	\end{minipage}
	\label{cifar_10}
}%

\subfigure[ImageNet-Subset (5 tasks)]{
	\begin{minipage}[t]{0.24\linewidth}
		\centering
		\includegraphics[width=1.6in]{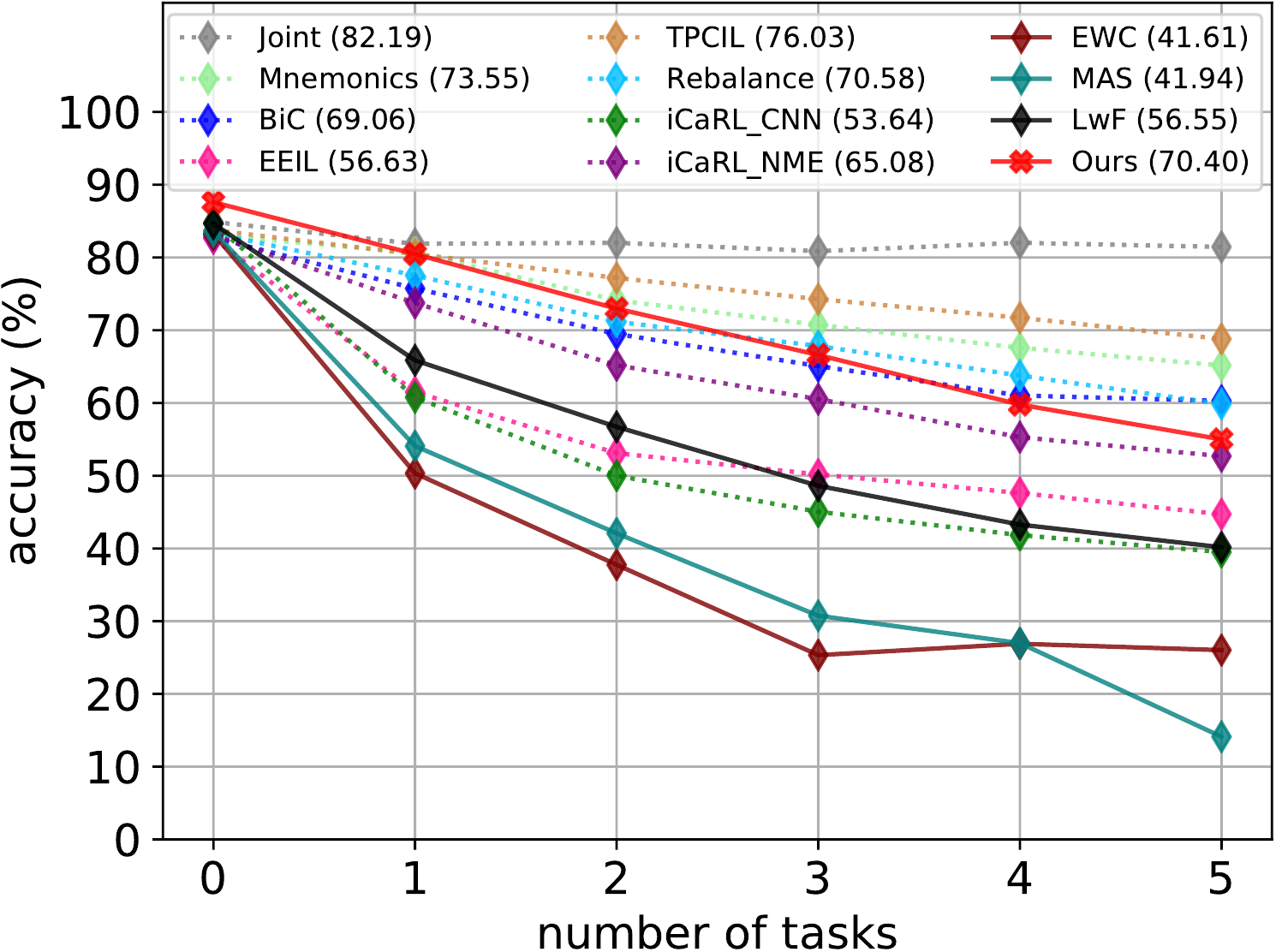}
		%\caption{fig1}
	\end{minipage}%
	\label{imagenet_5}
}%
\subfigure[ImageNet-Subset (10 tasks)]{
	\begin{minipage}[t]{0.24\linewidth}
		\centering
		\includegraphics[width=1.6in]{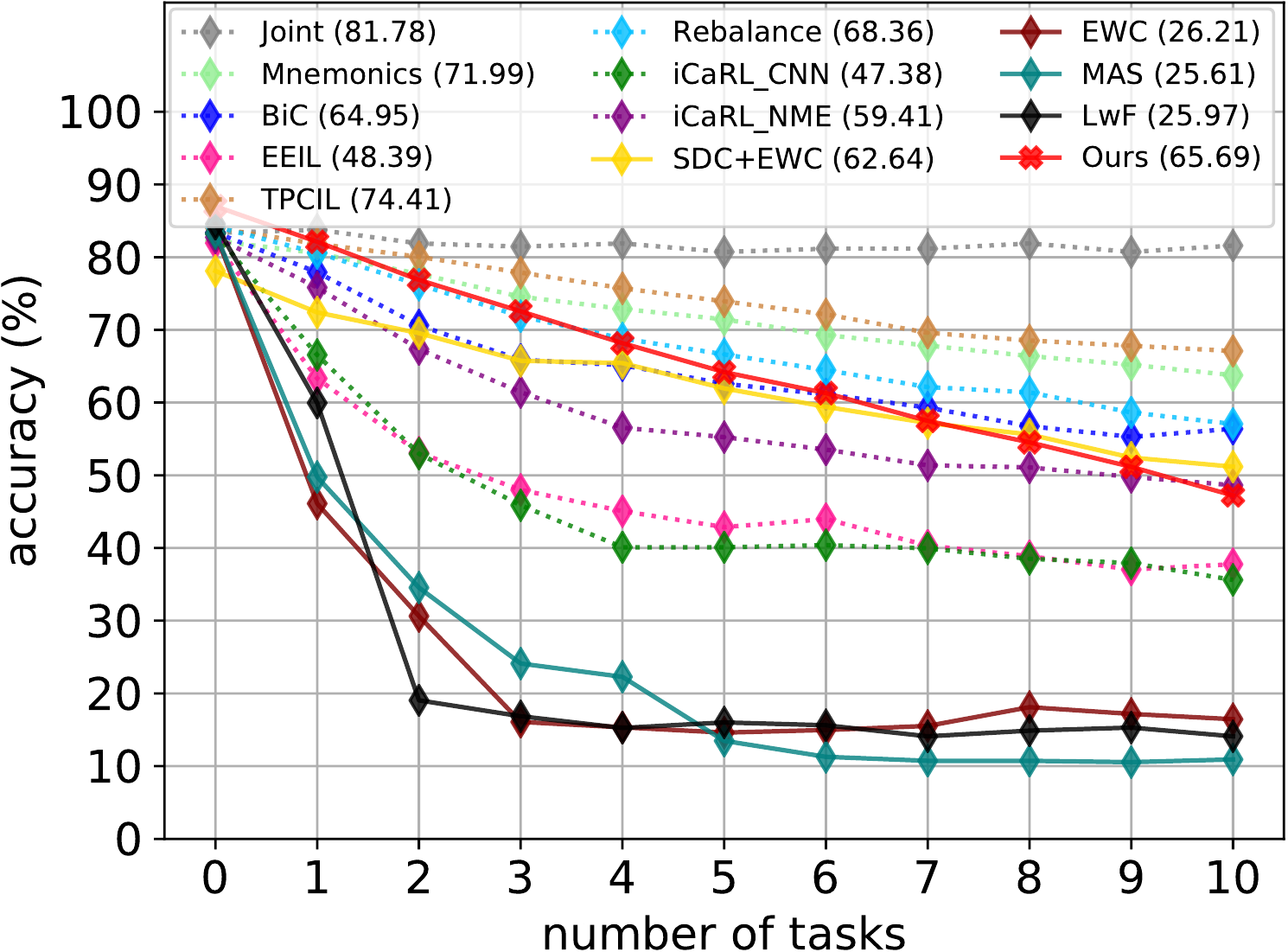}
		%\caption{fig2}
	\end{minipage}%
	\label{imagenet_10}
}%
\subfigure[Caltech-101 (10 tasks)]{
	\begin{minipage}[t]{0.24\linewidth}
		\centering
		\includegraphics[width=1.6in]{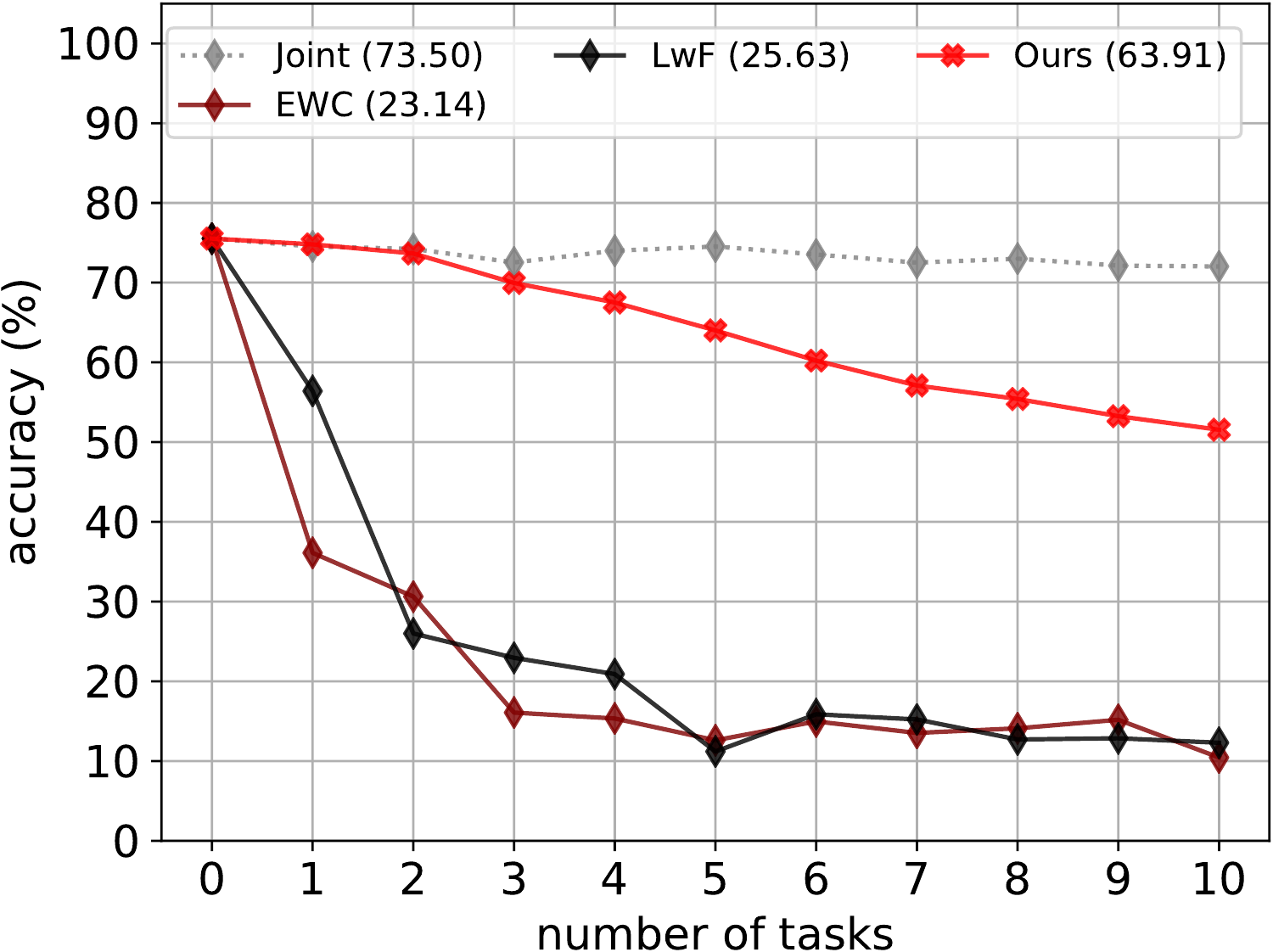}
		%\caption{fig2}
	\end{minipage}
	\label{caltech_10}
}%
\subfigure[Flowers-102 (10 tasks)]{
	\begin{minipage}[t]{0.24\linewidth}
		\centering
		\includegraphics[width=1.6in]{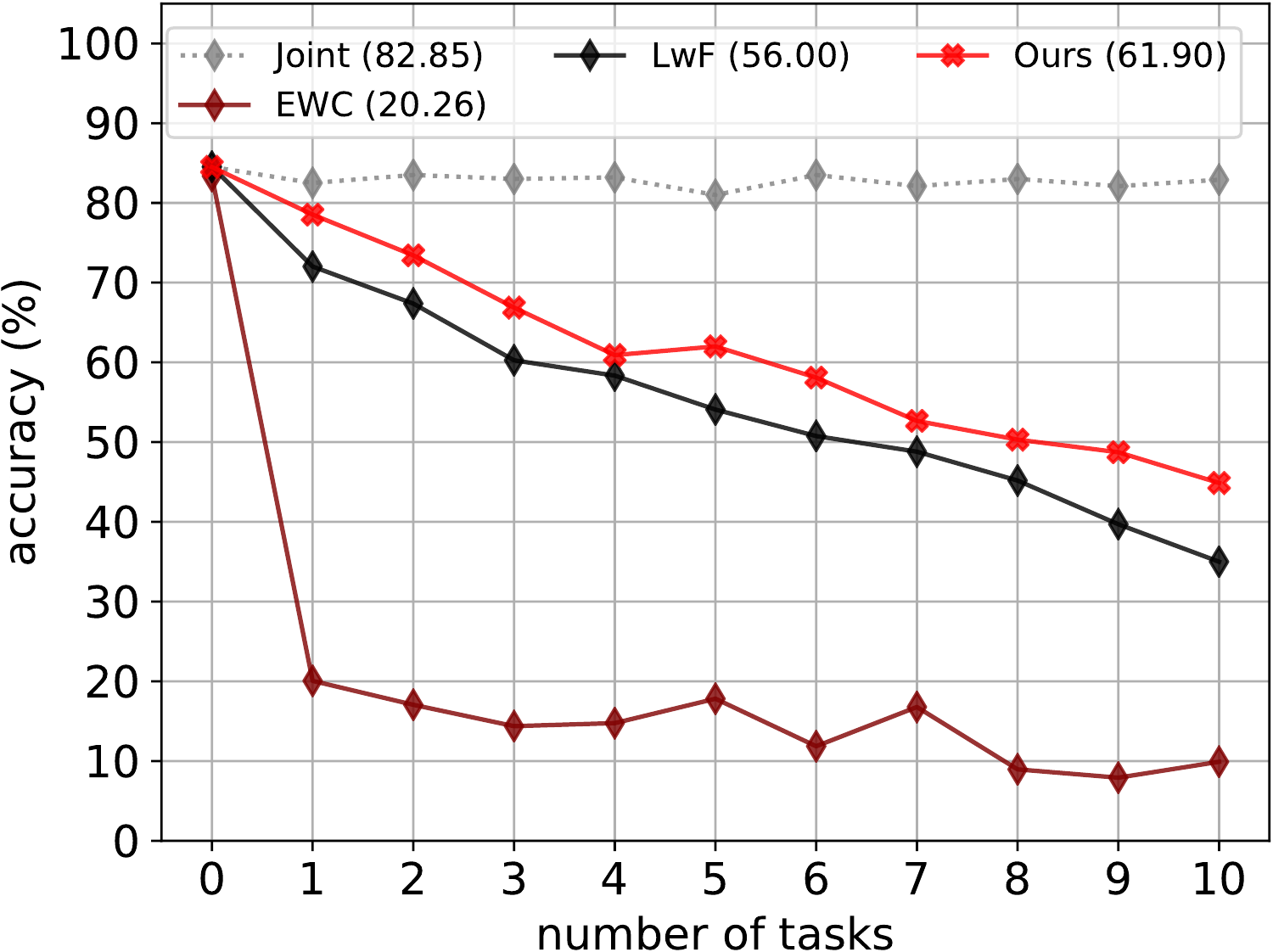}
		%\caption{fig2}
	\end{minipage}
	\label{flower_10}
}%
\centering
\caption{The incremental learning performance of the proposed method on four datasets with different task configurations. Solid lines indicate \textbf{exemplar-based} methods, and dashed lines indicate \textbf{exemplar-free} methods.}
\label{allresults}
\end{figure*}

In this section, we compare the proposed method with existing exemplar-free methods~\cite{Aljundi2018, Kirkpatrick2017, Li2018, Yu2020} as well as some excellent exemplar-based methods~\cite{castro2018end, Hou2019, Liu2020, Rebuffi2017, tao2020topology, wu2019large}, which adopt the same experimental setting as ours. In all experiments, we follow the experimental configuration of~\cite{Hou2019}. That is, we start from a model pre-trained on half of the total classes for all datasets. The top-1 accuracy after learning each task and the average accuracy over all incremental tasks are reported in Fig.~\ref{allresults}.

\textbf{On CIFAR-100}, we divide the remaining half of the classes into 1, 2, 5 and 10 tasks. As shown in Fig.~\ref{cifar_1} -- Fig.~\ref{cifar_10}, the performance of the proposed method surpasses all exemplar-free incremental learning methods by a large margin. Compared with the current state-of-the-art exemplar-free method SDC+EWC~\cite{Yu2020}, our method boosts the average accuracy by 5.75\% and 3.53\% in the 5-task and 10-task settings respectively. Besides, the performance of our base model will not be sacrificed, which is the weakness of the embedding network. Compared with rehearsal-based methods which make use of exemplars, our method surpasses iCaRL with NME~\cite{Rebuffi2017}, EEIL and BiC, and achieves very close performance to LUCIR~\cite{Hou2019}.

\textbf{On ImageNet-Subset}, the conclusion is consistent with that of CIFAR-100. The performance of our method surpasses all exemplar-free methods by a large margin, and is close to one of the best exemplar-based methods, LUCIR~\cite{Hou2019}. As shown in Fig.~\ref{imagenet_5} -- Fig.~\ref{imagenet_10}, in the first few tasks, the performance of our method even surpasses LUCIR. In terms of average accuracy, we achieve almost the same performance as LUCIR in the 5-task setting, and the performance gap is less than 3\% in the 10-task setting. It is worth noting that to our best knowledge, the proposed method is the first to successfully train a generative network for knowledge transfer and maintenance on high-resolution datasets like ImageNet in a totally data-free manner.

\textbf{On Caltech-101 and Flowers-102}, we conduct the 10-task learning experiment. As shown in Fig.~\ref{caltech_10} and Fig.~\ref{flower_10}, our method also works well and its performance is much better than LwF and EWC. In particular, the Flowers-102 dataset is designed for fine-grained classification, and the proposed method can handle this kind of tasks as well.

\section{Discussion}
In reality, data privacy is often of high concern, which makes challenging exemplar-free incremental learning matter. Reducing the natural performance gap between exemplar-free and exemplar-based methods deserves more attention in the community. In this paper, we innovatively introduce a generative model as a delegator of knowledge transfer. It is effectively trained with the goal of transferring the knowledge of a pre-trained classification model to a randomly re-initialized new model. Then the knowledge transfer capability of the delegator is exploited for maintenance of knowledge in the context of CIL. Based on this, we achieve excellent performance, even surpassing some exemplar-based methods. The proposed idea may have some potential applications such as knowledge amalgamation and unsupervised domain adaptation. For limitations, as shown in Fig.~\ref{allresults}, it tends to decline rapidly as the number of incremental tasks increases, especially on difficult datasets like ImageNet-Subset. This may be due to the increased difficulty in training SKD to extract the knowledge of the model after many rounds of incremental learning. We will explore to solve these problems in the future.

\section{Conclusions}
Consolidating the performance of the model on old tasks without accessing any old task data is a quite challenging problem in CIL. In this paper, we proposed a SKD, which can absorb the knowledge of the model on old tasks, and reproduce the feature extraction capability from scratch by constructing informative data. With the help of the novel imitate \& explore adversarial training strategy, the pseudo data exhibit similar distribution to real data in the feature space from the perspective of the pre-trained model. Based on the strong knowledge transfer and maintenance capability achieved by SKD, our incremental learning method surpasses all existing exemplar-free methods and some excellent exemplar-based methods. The novel idea of attacking exemplar-free incremental learning from the perspective of knowledge transfer is promising and inspiring.

\bibliographystyle{IEEEtran}
\bibliography{main}{}

\clearpage
\appendix
\subsection{Detailed Architectures of SKD}
\begin{table}[t]
  \caption{Architecture of SKD for low-resolution datasets.}
  \label{table_small}%
  \begin{center}
    \begin{tabular}{c}
    \toprule
    FC, Reshape, BN \\
    \midrule
    Upsample $2\times$ \\
    \midrule
     Conv (3 $\times$ 3 $\times$ 128), BN, LeakyReLU \\
    \midrule
    Upsample $2\times$ \\
    \midrule
     Conv (3 $\times$ 3 $\times$ 64), BN, LeakyReLU \\
    \midrule
     Conv (3 $\times$ 3 $\times$ 3), Tanh \\
    \midrule
    BN \\
    \bottomrule
	\end{tabular}%
  \end{center}
\end{table}%

\begin{table}[t]
  \caption{Architecture of SKD for high-resolution datasets.}
  \label{table_big}%
  \begin{center}
    \begin{tabular}{c}
    \toprule
    FC, Reshape, BN \\
    \midrule
    Deconv (3 $\times$ 3 $\times$ 512) 2$\times$ , BN, LeakyReLU \\
    \midrule
    Deconv (3 $\times$ 3 $\times$ 256) 2$\times$ , BN, LeakyReLU \\
    \midrule
    Deconv (3 $\times$ 3 $\times$ 128) 2$\times$ , BN, LeakyReLU \\
    \midrule
    Deconv (3 $\times$ 3 $\times$ 64) 2$\times$ , BN, LeakyReLU \\
    \midrule
     Conv (3 $\times$ 3 $\times$ 3), Tanh \\
    \midrule
    BN \\
    \bottomrule
    \end{tabular}%
  \end{center}
\end{table}%

Table~\ref{table_small} and Table~\ref{table_big} list the detailed architectures of the proposed SKD. For low-resolution datasets such as CIFAR-100, the structure in Table~\ref{table_small} is adopted in the paper. For the other three high-resolution datasets, namely ImageNet-Subset, Caltech-101 and Flowers-102, the structure in Table~\ref{table_big} is adopted.

\subsection{More Results on ImageNet-Subset, Caltech-101 and Flowers-102}
\begin{figure}[t]
\centering
\includegraphics[width=0.9\columnwidth]{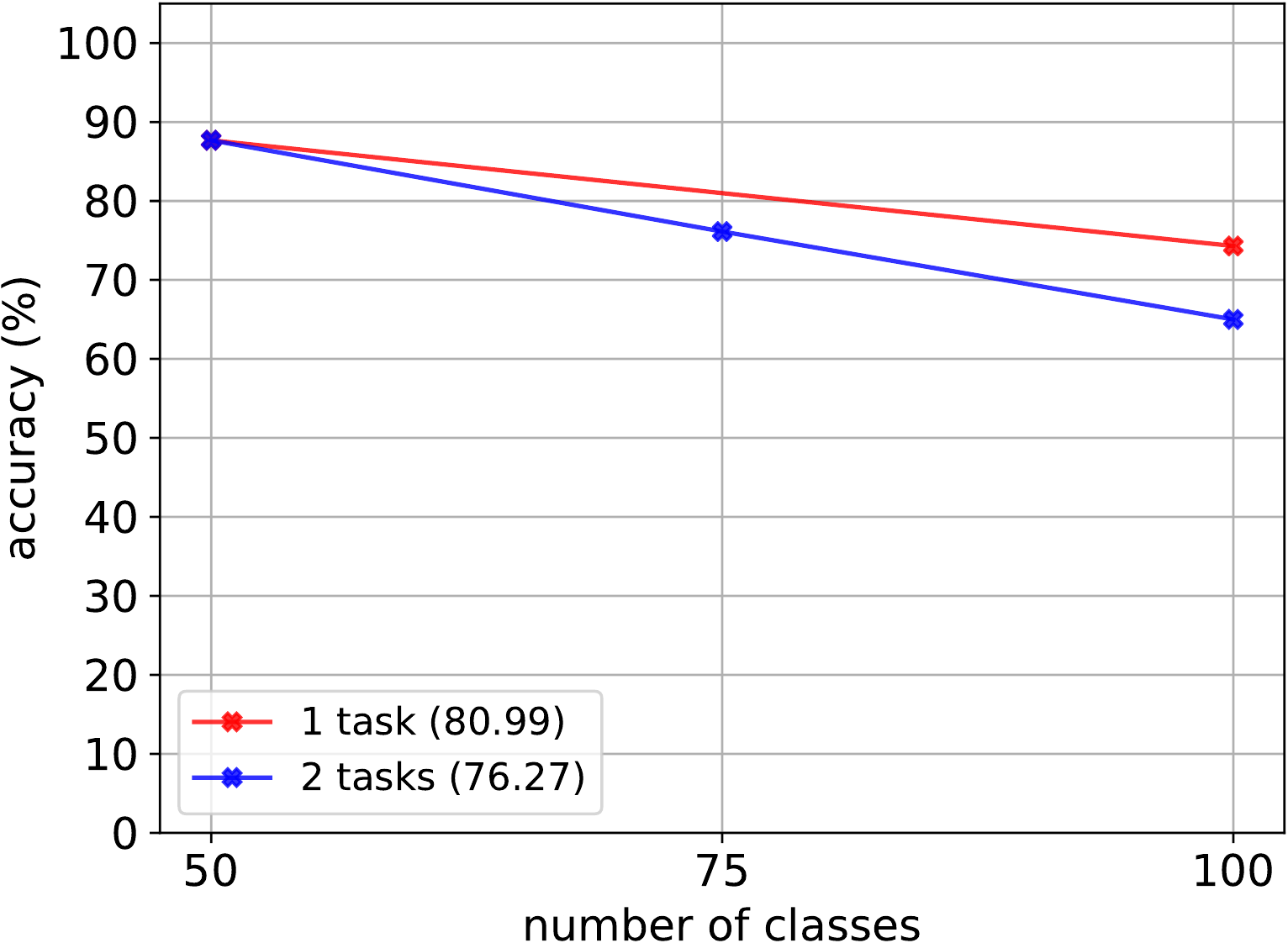}
  \caption{Results of the 1-task and 2-task settings on ImageNet-Subset.}
\label{imgenet_ots}
\end{figure}

\begin{figure}[t]
\centering
\includegraphics[width=0.9\columnwidth]{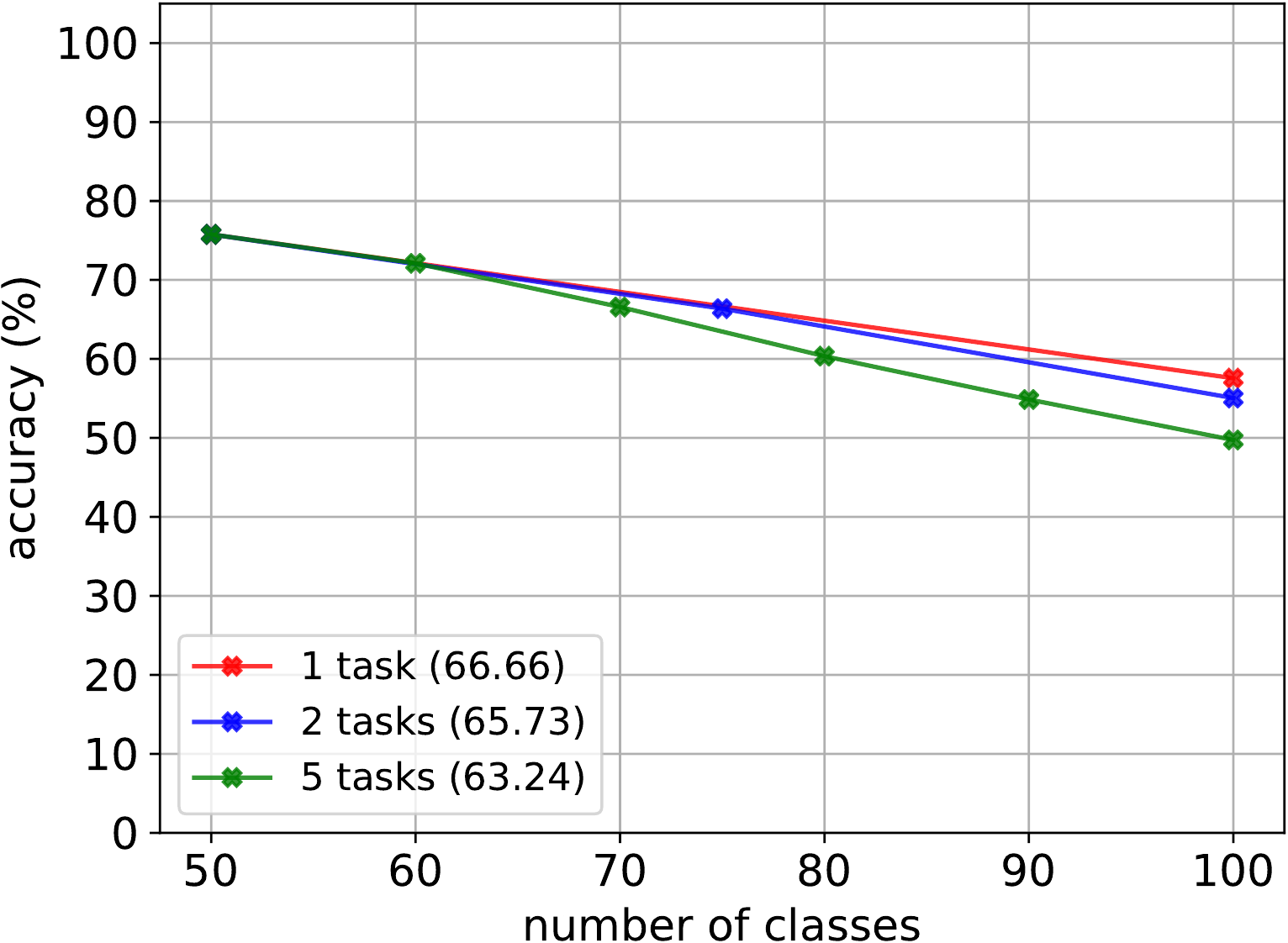}
  \caption{Results of the 1-task, 2-task and 5-task settings on Caltech-101.}
\label{clatech_ots}
\end{figure}

\begin{figure}[t]
\centering
\includegraphics[width=0.9\columnwidth]{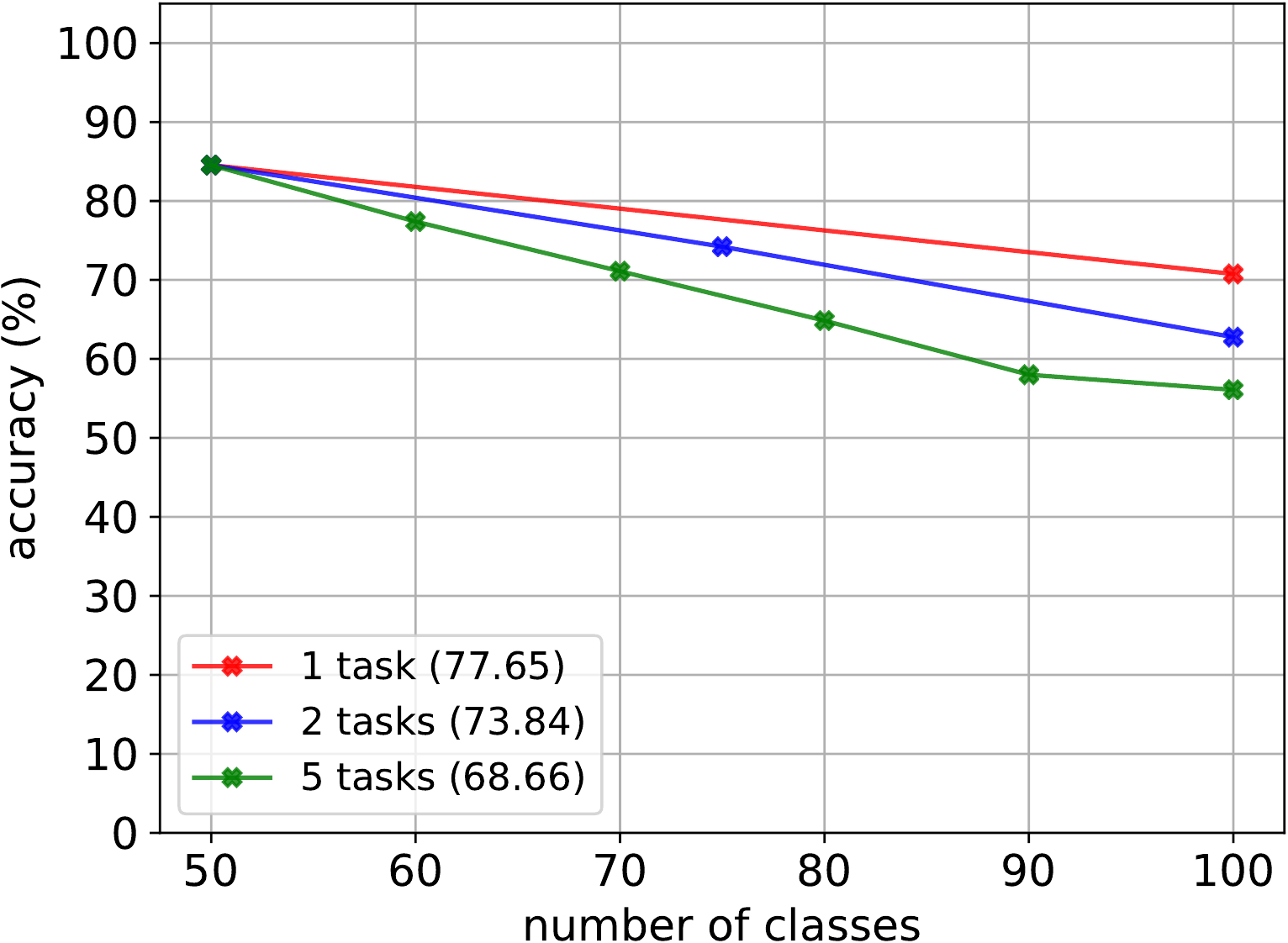}
  \caption{Results of the 1-task, 2-task and 5-task settings on Flowers-102.}
\label{flowers_ots}
\end{figure}

We also did extensive experiments on the three datasets, \ie ImageNet-Subset, Caltech-101 and Flowers-102. Due to lack of a baseline model for comparison, we did not paste the results into the main content. Fig.~\ref{imgenet_ots}, Fig.~\ref{clatech_ots} and Fig.~\ref{flowers_ots} show supplementary results on these datasets.

\subsection{Visualization of SKD Generated Data}
\begin{figure*}[t]
\centering
\subfigure[CIFAR-100]{
	\begin{minipage}[t]{0.5\linewidth}
		\centering
		\includegraphics[width=3in]{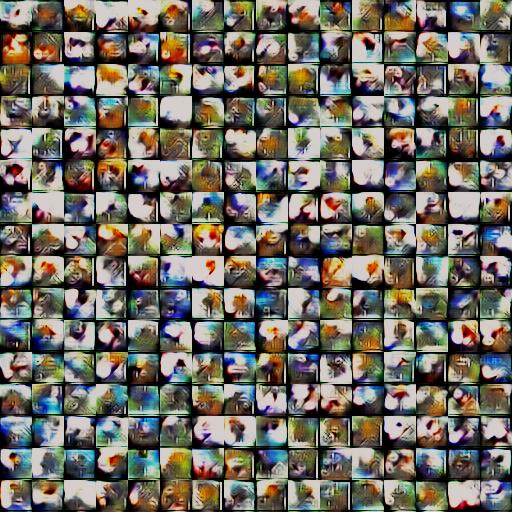}
		%\caption{fig1}
	\end{minipage}%
	\label{vis_cifar}
}%
\subfigure[Caltech-101]{
	\begin{minipage}[t]{0.5\linewidth}
		\centering
		\includegraphics[width=3in]{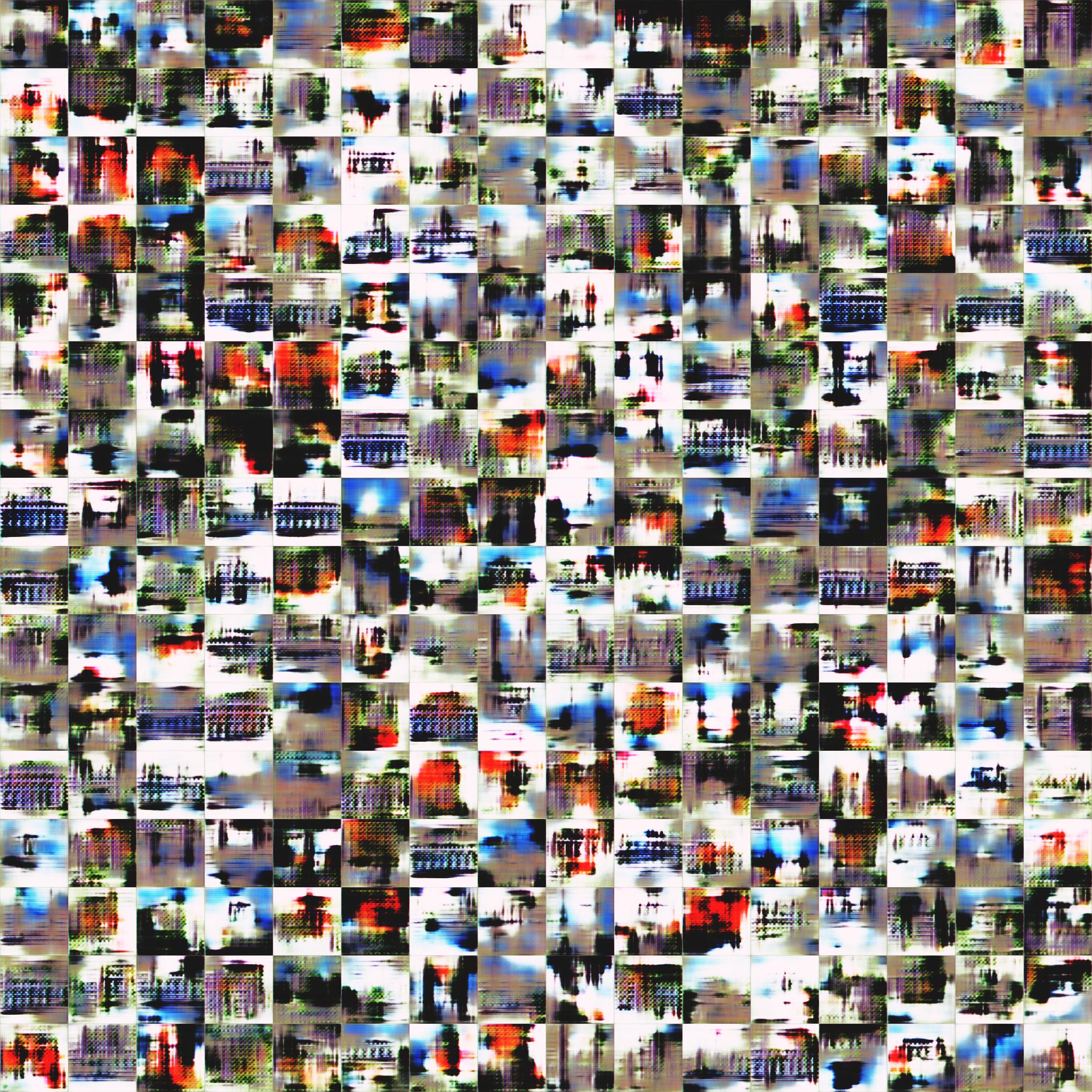}
		%\caption{fig2}
	\end{minipage}%
	\label{vis_clatech}
}%

\subfigure[Flowers-102]{
	\begin{minipage}[t]{0.5\linewidth}
		\centering
		\includegraphics[width=3in]{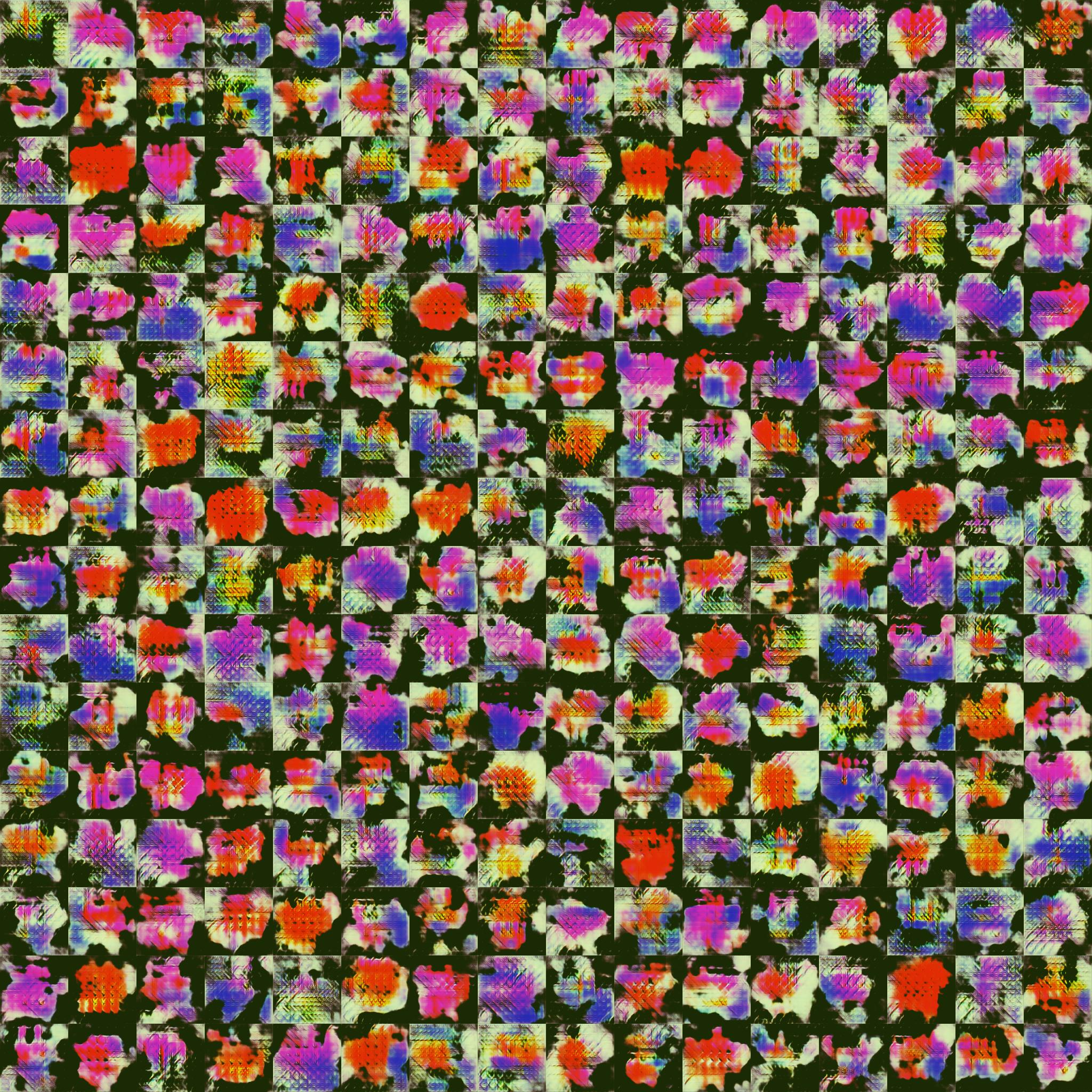}
		%\caption{fig1}
	\end{minipage}%
	\label{vis_flowers}
}%
\subfigure[Imagenet-Subset]{
	\begin{minipage}[t]{0.5\linewidth}
		\centering
		\includegraphics[width=3in]{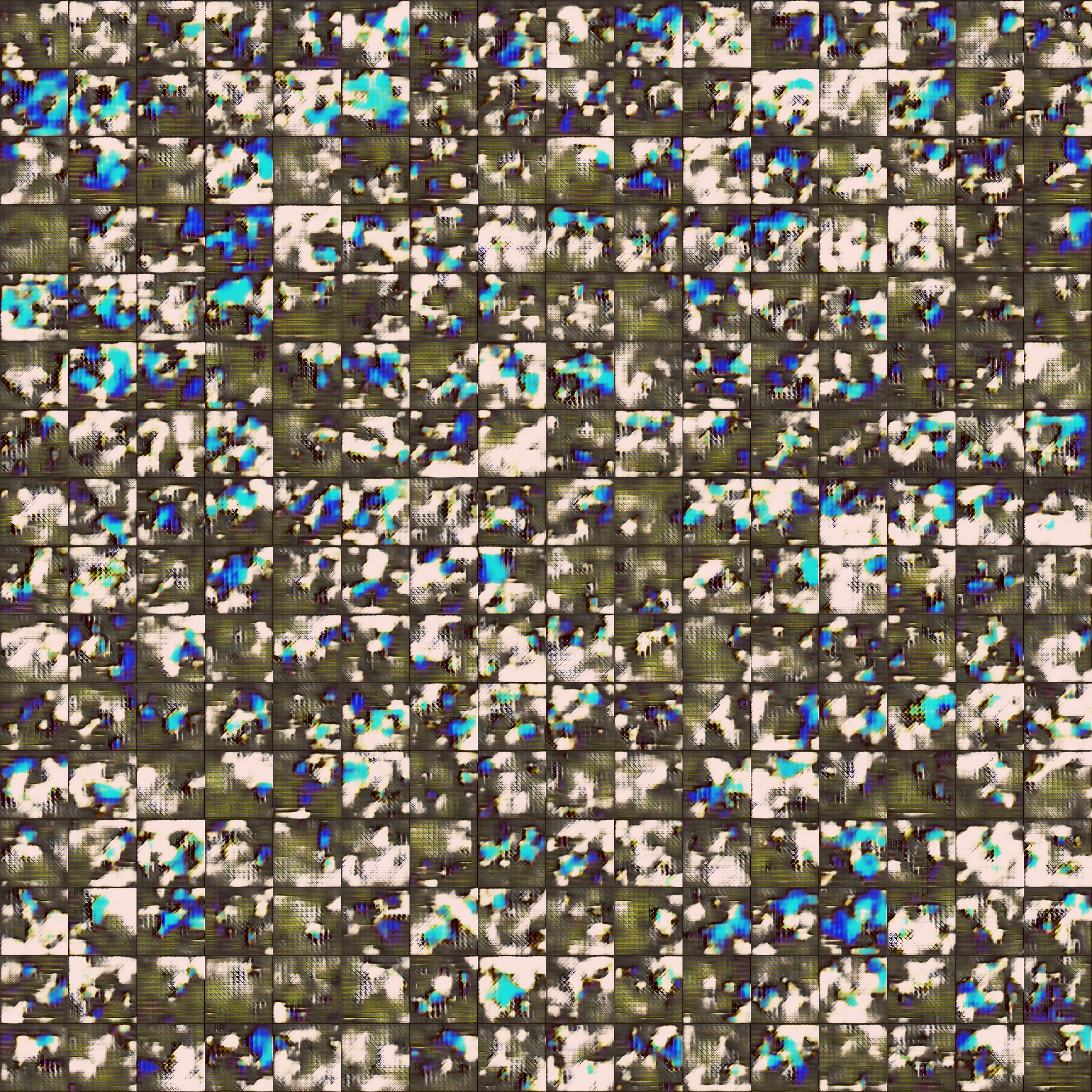}
		%\caption{fig2}
	\end{minipage}%
	\label{vis_imagenet}
}%
\centering
\caption{Visualization of the pseudo samples generated by SKD on the four datasets.}
\label{vis_all}
\end{figure*}
We also did some exploration on what SKD can generate. Some pseudo data generated by SKD are visualized in Fig.~\ref{vis_all}. Notably, the actual size of the generated images is $32 \times 32$ for CIFAR-100, $128 \times 128$ for Caltech-101 and Flowers-102, and $224 \times 224$ for ImageNet-Subset. Readers can see the details by zooming in the images.
\end{document}